\documentclass[runningheads]{llncs}
\usepackage{graphicx}
\usepackage{amsmath,amssymb} 
\usepackage{color}
\usepackage{xspace}

\newcommand{\M}[1]{\mathtt{#1}}
\newcommand{\V}[1]{\M{#1}}

\newcommand{\arr}[2]{\begin{array}{#1} #2\end{array}}
\newcommand{\mat}[2]{\left[\!\!\arr{#1}{#2}\!\!\right]}

\def\gb{Gr{\"o}bner basis\xspace}

\def \solverCv{$\text{R6P}_{\V{v},\V{C}}^{\V{w},\V{t}}$\xspace}
\def \solverw{ $\text{R6P}_{\V{v},\V{C},\V{t}}^{\V{w}}$\xspace} 
\def \solverwt{$\text{R6P}_{\V{v},\V{C},\V{t}}^{\V{w},\V{t}}$\xspace}
\def\solverall{$\text{R6P}_{\V{v},\V{C},\V{w},\V{t}}^{[\V{v}]_\times}$\xspace}

\begin{document}

\title{Linear solution to the minimal absolute pose rolling shutter problem\thanks{This work was supported by the European Regional Development Fund under the project IMPACT (reg. no. CZ.02.1.01/0.0/0.0/15\_003/0000468), EC H2020-ICT-731970 LADIO project, ESI Fund, OP RDE programme under the project International Mobility of Researchers MSCA-IF at CTU No. CZ.02.2.69/0.0/0.0/17\_050/0008025, and Grant-in-Aid for Scientific Research (Grant No. 16H02851) of the Ministry of Education, Culture, Sports, Science and Technology of Japan. A part of this work was done when Zuzana Kukelova was visiting the National Institute of Informatics (NII), Japan, funded in part by the NII MOU grant.}} 
\titlerunning{Linear RS absolute pose} 


\author{Zuzana Kukelova\textsuperscript{1}\quad\quad Cenek Albl\textsuperscript{2}\quad\quad Akihiro Sugimoto\textsuperscript{3}\quad\quad Tomas Pajdla\textsuperscript{2}}
\institute{ Visual Recognition Group (VRG), FEE, CTU in Prague\textsuperscript{1} \\ Czech Institute of Informatics, Robotics and Cybernetics (CIIRC), CTU in Prague \textsuperscript{2} \\ National Institute of Informatics, 
Tokyo, Japan \textsuperscript{3}}

%

\authorrunning{Z. Kukelova et al.} 


\maketitle

\begin{abstract}
This paper presents new efficient solutions to the rolling shutter camera absolute pose problem. Unlike the state-of-the-art polynomial solvers, we approach the problem using simple and fast linear solvers in an iterative scheme. We present several solutions based on fixing different sets of variables and investigate the performance of them thoroughly. We design a new alternation strategy that estimates all parameters in each iteration linearly by fixing just the non-linear terms. Our best 6-point solver, based on the new alternation technique, shows an identical or even better performance than the state-of-the-art R6P solver and is two orders of magnitude faster. In addition, a linear non-iterative solver is presented that requires a non-minimal number of 9 correspondences but provides even better results than the state-of-the-art R6P.  Moreover, all proposed linear solvers provide a single solution while the state-of-the-art R6P provides up to 20 solutions which have to be pruned by expensive verification.

\keywords{Rolling shutter  \and Absolute pose \and Minimal solvers}
\end{abstract}

\section{Introduction}

Rolling shutter (RS) cameras are omnipresent. They can be found in smartphones, consumer, professional, and action cameras and even in self-driving cars. RS cameras are cheaper, and easier to produce, than global shutter cameras. They also posses other advantages over the global shutter cameras, such as higher achievable frame-rate or longer exposure times. 

There is, however, a significant drawback when using them for computer vision applications. When the scene or the camera is moving during image capture, images produced by RS cameras will become distorted. The amount and type of distortion depends on the type and speed of camera motion and on the depth of the scene. 
It has been shown that RS image distortions can cause problems for standard computer vision methods such as Structure from Motion~\cite{Hedborg2012}, visual SLAM~\cite{Klein2009} or multi-view dense stereo~\cite{Saurer2013}. Therefore, having a special camera model for rolling shutter cameras is desirable.

The camera absolute pose computation is a fundamental problem in many computer vision tasks such as Structure from Motion, augmented reality, visual SLAM, and visual localization. The problem is to compute the camera pose from 3D points in the world and their 2D projections into an image. The minimal number of correspondences necessary to solve the absolute pose problem for a perspective calibrated camera is three. The first solution to this problem was introduced by Grunert~\cite{Grunert-1841} and since then it was many times revisited~\cite{Haralick1991,Ameller02camerapose,Fischler-Bolles-ACM-1981}. Other work has focused on computing the absolute pose from a larger than the minimal number of correspondences~\cite{Lepetit-IJCV-2009,Quan1999,Triggs1999,Wu2006,Zhi2002}. All of the previous work consider a perspective camera model, which is not suitable for dynamic RS cameras.

Recently, as RS cameras became more and more common, the focus turned to computing camera absolute pose from images containing RS effects. First, several RS camera motion models were introduced in~\cite{Meingast2005}. A solution to RS absolute pose using non-minimal (eight and half) number of points was presented in~\cite{Aitaider2006}. It relied on a non-linear optimization and required a planar scene. 

In~\cite{Hedborg2011}, video sequences were exploited and the absolute camera pose was computed sequentially using a non-linear optimization starting from the previous camera pose. Another approach using video sequences was used for visual SLAM in~\cite{Klein2009} where the camera motion estimated from previous frames was used to compensate the RS distortion in the next frame prior to the optimization.

A polynomial solution that is globally optimal was presented in~\cite{Magerand2012}. It uses Gloptipoly~\cite{Henrion2009} solver to find a solution from 7 or more points. Authors show that the method provides better results than~\cite{Aitaider2006}, but the runtime is in the order of seconds, making it impractical for typical applications such as RANSAC. 

The first minimal solution to the rolling shutter camera absolute pose problem was presented in~\cite{Albl-CVPR-2015}. It uses the minimal number of six 2D to 3D point correspondences and the \gb{} method to generate an efficient solver. The proposed R6P is based on the constant linear and angular velocity model as in~\cite{Aitaider2006,Magerand2012,Hedborg2012} but it uses the first order approximation to both the camera orientation and angular velocity, and, therefore, it requires an initialization of the camera orientation, e.g., from P3P~\cite{Fischler-Bolles-ACM-1981}. Paper~\cite{Albl-CVPR-2015} has shown that R6P solver significantly outperforms perspective P3P solver in terms of camera pose precision and the number of inliers captured in the RANSAC loop. 

\subsection{Motivation}
It has been demonstrated in the literature that RS camera absolute pose is beneficial and often necessary when dealing with RS images from moving camera or dynamic scene. Still, until now, all the presented solutions have significant drawbacks that make them impractical for general use. 

The state-of-the-art solutions require a non-minimal or a larger number of points~\cite{Aitaider2006,Magerand2012}, planar scene~\cite{Aitaider2006}, video sequences~\cite{Hedborg2011,Hedborg2012,Klein2009}, are very slow~\cite{Magerand2012} and provide too many solutions~\cite{Albl-CVPR-2015}. 

If one requires a practical algorithm similar to P3P, but working on RS images, the closest method available is R6P~\cite{Albl-CVPR-2015}. However, R6P still needs around $1.7ms$ to compute the camera pose, compared to around $3\mu s$ for P3P. Therefore, in typical applications where P3P is used, one would suffer a several orders of magnitude slowdown compared to P3P. This makes it hard to use for real-time applications such as augmented reality. In addition, R6P provides up to 20 real solutions, which need to be verified. This makes tasks like RANSAC, which uses hundreds or thousands of iterations and verifies all solutions, extremely slow compared to P3P. This motivates us to create a solution to RS absolute pose problem with similar performance to R6P~\cite{Albl-CVPR-2015} and  runtime comparable to P3P.

\subsection{Contribution}
In this work we present solutions that remove previously mentioned drawbacks of the state-of-the-art methods and provide practical and fast rolling shutter camera absolute pose solvers. We take a different approach to formulating the problem and propose linear solutions to rolling shutter camera absolute pose.
Specifically, we present the following RS absolute camera pose solvers:
\begin{itemize}
    \item a 6-point linear iterative solver, which provides identical or even better solutions than R6P in $10\mu s$, which is up to $170\times$ faster than R6P. This solver is based on a new alternating method;
    \item two 6-point linear iterative solvers that outperform R6P for purely translational motion;
    \item a 9-point linear non-iterative solver that provides more accurate camera pose estimates than R6P in $20\mu s$;
\end{itemize}

 All solvers are easy to implement and they return a single solution. We formulate the problem of RS camera absolute pose in Section~\ref{sec:formulation}. Derivations of all new solvers are in Section~\ref{sec:lin_solvers}. Section~\ref{sec:experiments} contains experiments verifying the feasibility of the proposed solvers and it compares them against P3P and R6P~\cite{Albl-CVPR-2015}.
\section{Problem formulation}
\label{sec:formulation}
For calibrated perspective cameras, the projection equation can be written as 
\begin{equation}
\lambda_i \V{x}_i = \M{R}\V{X}_i+\V{C},
\label{eq:persp_proj}
\end{equation}
where $\M{R}$ and $\V{C}$ are the rotation and translation bringing a 3D point $\V{X}_i$ from a world coordinate system to the camera coordinate system with\, $\V{x}_i = \left[r_i,c_i,1\right]^{\top}$, and scalar $\lambda_i \in \mathbb{R}$. 
For RS cameras, every image row is captured at different time and hence at a different position when the camera is moving during the image capture. Camera rotation $\M{R}$ and translation $\V{C}$ are therefore 
functions of the image row $r_i$ being captured
\begin{equation}
\lambda_i \V{x}_i=  \lambda_i \mat{l}{r_i\\c_i\\1} = \M{R}(r_i)\V{X}_i+\V{C}(r_i).
\label{eq:proj_rs}
\end{equation}

In recent work~\cite{Albl-CVPR-2015,Saurer2013,Magerand2012,Meingast2005,Hedborg2012,Aitaider2006}, it was shown that 
for the short time-span of a frame capture, the camera translation $\V{C}(r_i)$  can be approximated with a simple constant velocity model as
\begin{equation}
\V{C}(r_i) = \V{C} + (r_i-r_0)\V{t},
\label{eq:C_ri}
\end{equation}
where $\V{C}$ is the camera center corresponding to the perspective case, i.e. when $r_i=r_0$, and $\V{t}$ is the translational velocity.

The camera rotation $\M{R}(r_i)$ can be decomposed into two rotations to represent the camera initial orientation by $\M{R}_{\V{v}}$ and the change of orientation during frame capture by $\M{R}_{\V{w}}(r_i-r_0)$. 

In~\cite{Magerand2012,Albl-CVPR-2015}, it was observed that it is usually sufficient to linearize $\M{R}_{\V{w}}(r_i-r_0)$ around the initial rotation 
$\M{R}_\V{v}$
using the first order Taylor expansion such that
\begin{equation}
\lambda_i \mat{l}{r_i\\c_i\\1} = \left(\M{I}+(r_i-r_0)[\V{w}]_\times\right)\M{R}_\V{v}\V{X}_i +\V{C}+(r_i-r_0)\V{t},
\label{eq:model_lin}
\end{equation}
where $[\V{w}]_\times$ is a skew-symmetric matrix of vector $\V{w}$.
The model~(\ref{eq:model_lin}), with linearized rolling shutter rotation, will deviate from the reality with increasing rolling shutter effect. Still, it is usually sufficient for most of the rolling shutter effects present in real situations.

In~\cite{Albl-CVPR-2015}, a linear approximation to the camera orientation $\M{R}_\V{v}$ was used to solve the rolling shutter absolute pose problem from a minimal number of six  2D-3D point correspondences.
This model has the form
\begin{equation}
\lambda_i \mat{l}{r_i\\c_i\\1} = \left(\M{I}+(r_i-r_0)[\V{w}]_\times\right)\left(\M{I}+[\V{v}]_\times\right)\V{X}_i +\V{C}+(r_i-r_0)\V{t}.
\label{eq:model_double_lin}
\end{equation}
The drawback of the model~(\ref{eq:model_double_lin}) is that $\M{R}_{\V{v}}$ is often not small and thus cannot be linearized. Therefore, the accuracy of the model is dependent on the initial orientation of the camera in the world frame. In~\cite{Albl-CVPR-2015}, it was shown that the standard P3P algorithm~\cite{Fischler-Bolles-ACM-1981} is able to estimate camera orientation with sufficient precision even for high camera rotation velocity and therefore P3P can be used to bring the camera rotation matrix $\M{R}_\V{v}$ close to the identity, where~(\ref{eq:model_double_lin}) works reasonably.

The model~(\ref{eq:model_double_lin}) leads to a system of six quadratic
equations in six unknowns. This system has 20 solutions and it was solved in~\cite{Albl-CVPR-2015} using the \gb method~\cite{Cox-Little-etal-05,Kukelova-ECCV-2008}. The \gb solver~\cite{Albl-CVPR-2015} for the R6P rolling shutter problem requires the G-J elimination of a $ 196 \times 216$ matrix and computing the eigenvectors of a $20 \times 20$ matrix. The R6P solver runs for about 1.7ms and thus is too slow in many practical situations.

We will next show how to simplify this model by linearizing equation~(\ref{eq:model_double_lin}) and yet still obtaining a similar performance as the  \gb R6P absolute pose solver~\cite{Albl-CVPR-2015} for the original model~(\ref{eq:model_double_lin}). 

\section{Linear rolling shutter solvers}
\label{sec:lin_solvers}
We present here several linear iterative solvers to the minimal absolute pose rolling shutter problem. All these solvers start with the model~(\ref{eq:model_double_lin}) and they use six 2D-3D image point correspondences to estimate 12 unknowns $\V{v},\V{C},\V{w}$, and $\V{t}$. 
The proposed solvers differ in the way how the system~(\ref{eq:model_double_lin}) is linearized.
Additionally we propose a linear non-iterative 9 point absolute pose rolling shutter solver.

\subsection{$\textbf{R6P}_{\V{v},\V{C}}^{\V{w},\V{t}}$ solver}
\label{sec:solver1}
The \solverCv  solver is based on the idea of alternating between two linear solvers. The first $\text{R6P}_{\V{v},\V{C}}$ solver fixes the rolling shutter parameters $\V{w}$ and $\V{t}$ in~(\ref{eq:model_double_lin}) and estimates only the camera parameters $\V{v}$ and $\V{C}$.
The second $\text{R6P}_{\V{w},\V{t}}$ solver fixes the camera parameters $\V{v}$ and $\V{C}$ and estimates only the rolling shutter parameters $\V{w}$ and $\V{t}$. Both these partial solvers results in 12 linear equations in 6 unknowns that can be solved in the least square sense.  The motivation for this solver comes from the fact that even for larger rolling shutter speed, the camera parameters $\V{v}$ and $\V{C}$ can be estimated quite accurately.

The  \solverCv  solver starts with $\V{w}_0 = \V{0}$ and $\V{t}_0 = \V{0}$ and, in the first iteration, uses linear $\text{R6P}_{\V{v},\V{C}}$ solver to estimate $\V{v}_1$ and $\V{C}_1$. Using the estimated $\V{v}_1$ and $\V{C}_1$, the linear solver $\text{R6P}_{\V{w},\V{t}}$ estimates $\V{w}_1$ and $\V{t}_1$. This process is repeated until the desired precision is obtained or a maximum number of iterations is reached.

The \solverCv solver does not perform very well in our experiments, which we account to the fact that it never estimates the pose parameters $\V{v}$,$\V{C}$ and the motion parameters $\V{w}$,$\V{t}$ together in one step. Nevertheless, we present this solver as a logical first step when considering the iterative approach to RS absolute pose problem.
\subsection{$\textbf{R6P}_{\V{v},\V{C},\V{t}}^{\V{w}}$ and $\textbf{R6P}_{\V{v},\V{C},\V{t}}^{\V{w},\V{t}}$ solver}
\label{sec:solver2}
To avoid problems of the \solverCv solver, we introduce the \solverw solver. The \solverw solver alternates between two solvers, i.e.\ the linear $\text{R6P}_{\V{v},\V{C},\V{t}}$ solver, which fixes only the rolling shutter rotation $\V{w}$ and estimates $\V{v},\V{C}$ and $\V{t}$, and the $\text{R6P}_{\V{w}}$ solver that estimates only the rolling shutter rotation $\V{w}$ using the fixed  $\V{v},\V{C}$ and $\V{t}$. 
The $\text{R6P}_{\V{v},\V{C},\V{t}}$ solver solves 12 linear equations in 9 unknowns and the $\text{R6P}_{\V{w}}$ solver solves 12 linear equations in 3 unknowns in the least square sense.
Since the first $\text{R6P}_{\V{v},\V{C},\V{t}}$ solver assumes unknown rolling shutter translation, the camera parameters are estimated with better precision than in the case of the $\text{R6P}_{\V{v},\V{C}}$ solver.
Moreover, in many applications, e.g.\ cameras on a car, cameras often undergo only a translation motion, and therefore $\V{w}$ is negligible. In such situations, the first iteration of the $\text{R6P}_{\V{v},\V{C},\V{t}}$ solver already provides very precise estimates of the camera parameters.

Another approach is to use only the $\V{v}$ and $\V{C}$ estimated by $\text{R6P}_{\V{v},\V{C},\V{t}}$ solver and in the second step re-estimate the rolling shutter translation $\V{t}$ together with the rolling shutter rotation $\V{w}$ using the linear $\text{R6P}_{\V{w},\V{t}}$ solver. The solver based on this strategy will be referred to as~\solverwt{}.

The resulting solvers \solverw and \solverwt{}, again, alternate between the two linear solvers until the desired precision is obtained or a maximum number of iterations is reached. We show in the experiments that those solvers outperform R6P in the case of pure translational motion.

\subsection{$\textbf{R6P}_{\V{v},\V{C},\V{w},\V{t}}^{[\V{v}]_\times}$ solver}
The \solverall solver estimates all unknown parameters $\V{v},\V{C},\V{w}$ and $\V{t}$ together in one step. To avoid non-linearity  in~(\ref{eq:model_double_lin}), the solver fixes [$\V{v}]_\times$ that appears in the nonlinear term $[\V{w}]_\times [\V{v}]_\times$  in~(\ref{eq:model_double_lin}). Thus the solver solves equations
\begin{equation}
\lambda_i \mat{l}{r_i\\c_i\\1} = \left(\M{I}+(r_i-r_0)[\V{w}]_\times\right)\V{X}_i + [\V{v}]_\times\V{X}_i +  (r_i-r_0)[\V{w}]_\times[\V{\hat{v}}]_\times\V{X}_i +\V{C}+(r_i-r_0)\V{t},
\label{eq:model_double_lin2}
\end{equation}
where $\V{\hat{v}}$ is a fixed vector. 

 In the first iteration $\V{\hat{v}}$, is set to the zero vector and the term $(r_i-r_0)[\V{w}]_\times[\V{\hat{v}}]_\times\V{X}_i$ in~(\ref{eq:model_double_lin2}) disappears. This is usually a sufficient approximation. The explanation for this is as follows. After the initialization with P3P the camera rotation is already close to the identity and in real applications the rolling shutter rotation $\V{w}$ during the capture is usually small. Therefore, the nonlinear term $[\V{w}]_\times[\V{{v}}]_\times$ is small, sometimes even negligible, and thus it can be considered to be zero in the first iteration.   
 
In the remaining iterations we fix $\V{\hat{v}}$ in the  $(r_i-r_0)[\V{w}]_\times[\V{\hat{v}}]_\times\V{X}_i$  term to be equal to the $\V{v}_i$ estimated in the previous iteration of  the \solverall solver. Note that we fix only $\V{v}$ that appears in the nonlinear term $[\V{w}]_\times[\V{{v}}]_\times$ and there is still another term with $\V{v}$ in~(\ref{eq:model_double_lin2}) from which a new $\V{v}$ can be estimated.  Therefore, all parameters are estimated at each step which is a novel alternating strategy. To our knowledge, all existing algorithms that are based on the alternating optimization approach completely fix a subset of the variables, meaning that they cannot estimate all the variables in one step. 

The \solverall in each iteration solves only one system of 12 linear equations in 12 unknowns and is therefore very efficient. In experiments we will show that the~\solverall provides very precise estimates already after 1 iteration and after 5 iterations it has virtually the same performance as the state-of-the-art R6P solver~\cite{Albl-CVPR-2015}.
\subsection{R9P}
Our final solver is a non-iterative solver that uses a non-minimal number of nine 2D-3D point correspondences. We note that the projection equation~(\ref{eq:model_double_lin2}) can be rewritten as
\begin{equation}
\lambda_i \mat{l}{r_i\\c_i\\1} = \left(\M{I}+[\V{v}]_\times\right)\V{X}_i  +\V{C} +  (r_i-r_0)([\V{w}]_\times(\M{I}+[\V{{v}}]_\times)\V{X}_i + \V{t}).
\label{eq:model_double_lin3}
\end{equation}
We can substitute the term $[\V{w}]_\times(\M{I}+[\V{{v}}]_\times)$ in~(\ref{eq:model_double_lin3}) with a $3 \times 3$ unknown matrix $\M{R}_{\text{RS}}$.
After eliminating the scalar values $\lambda_i$ by multiplying equation~(\ref{eq:model_double_lin3}) from the left by the skew symmetric matrix for vector $\mat{ccc}{r_i & c_i & 1}^\top$
and without considering the internal structure of the matrix $\M{R}_{\text{RS}}$, we obtain three linear equations in 18 unknowns, i.e. $\V{v}, \V{C}, \V{t}$, and 9 unknowns in $\M{R}_{\text{RS}}$. Since only two from these tree equations are linearly independent we need nine 2D-3D point correspondences to solve this problem. 

Note that the original formulation~(\ref{eq:model_double_lin}) was an approximation to the real rolling shutter camera model and therefore the formulation with a general  $3 \times 3$ matrix $\M{R}_{\text{RS}}$ is yet a different approximation to this model.

\section{Experiments}
\label{sec:experiments}
We tested the proposed solvers on a variety of synthetic and real datasets and compared the results with the original R6P solver~\cite{Albl-CVPR-2015} as well as P3P. We followed the general pattern of experiments used in~\cite{Albl-CVPR-2015} in order to provide consistent comparison on the additional factor of experiments that are specific to our iterative solvers such as their convergence. 

To analyze the accuracy of the estimated camera poses and velocities, we used synthetic data in the following setup. A random set of 3D points was generated in a cubic region with $x,y,z\in[-1;1]$ and a camera with a distance $d\in[2;3]$ from the origin and pointing towards the 3D points. The camera was set to be calibrated, i.e. $\M{K}=\M{I}$ and the field of view was set to 45 degrees. Rolling shutter projections were created using a constant linear velocity and a constant angular velocity with various magnitudes. 

Using the constant angular velocity model for generating the data ensures that our data is not generated with the same model as the one that is estimated by the solvers (linear approximation to a rotation). Although the used model is just an approximation of the real rolling shutter model and we could have chosen another one, e.g. constant angular acceleration, we consider the constant angular velocity model as a reasonable description of the camera motion during the short time period of frame capture.

We used 6 points for the original R6P and all proposed R6P iterative solvers. In order to provide P3P with the same data, we used all possible triplets from the 6 points used by R6P and then chose the best result. For R9P we used 9 points. 
Unless stated otherwise, all iterative solvers were run for maximum 5 iterations in the experiments.
\subsection{Synthetic data}
\label{sec:exp_synth}
\begin{figure}[tb]
    \centering
    \includegraphics[width=0.45\columnwidth]{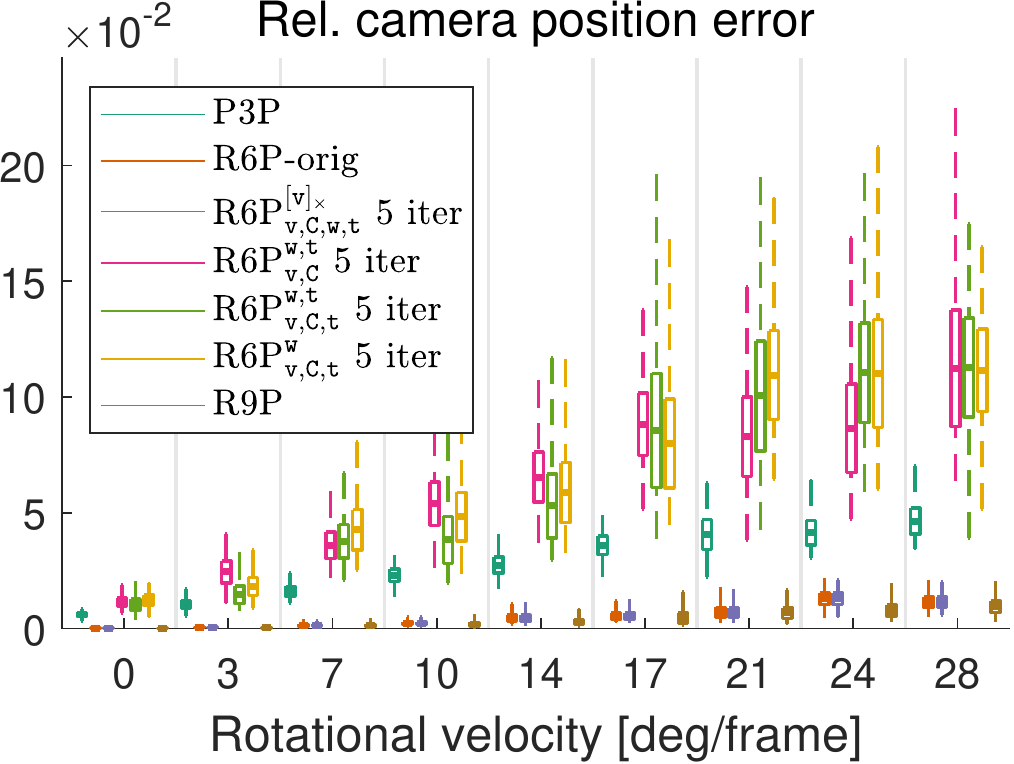}
      \hspace*{0.02\columnwidth}
    \includegraphics[width=0.45\columnwidth]{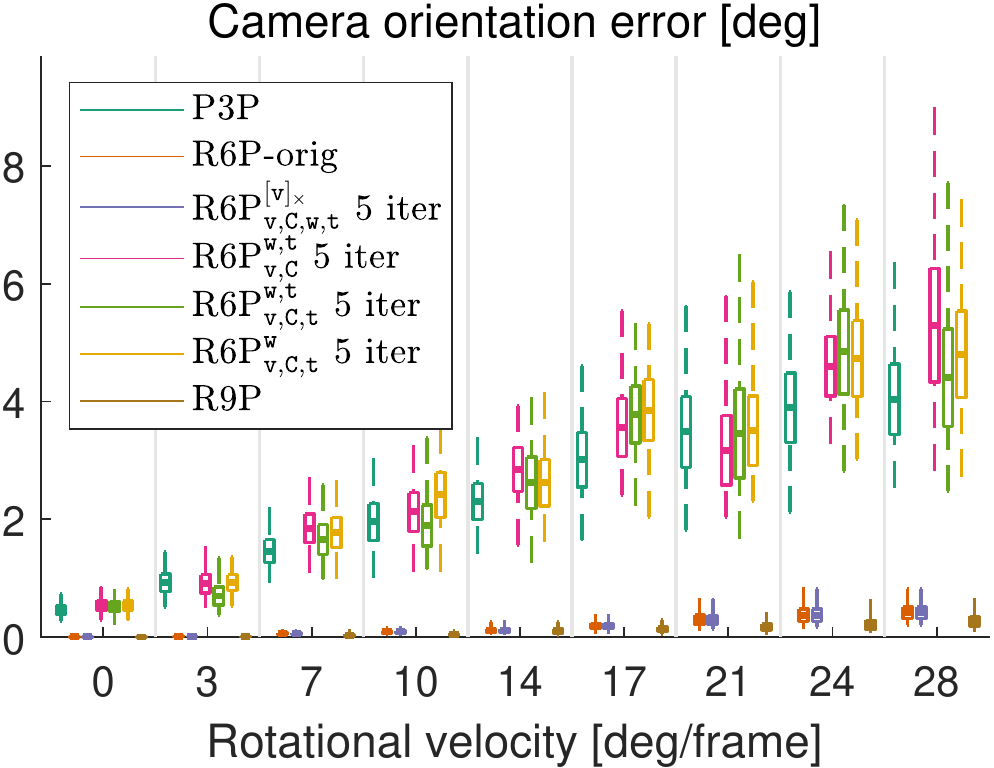} \\
    \includegraphics[trim={1cm 0 0 0},clip,width=0.45\columnwidth]{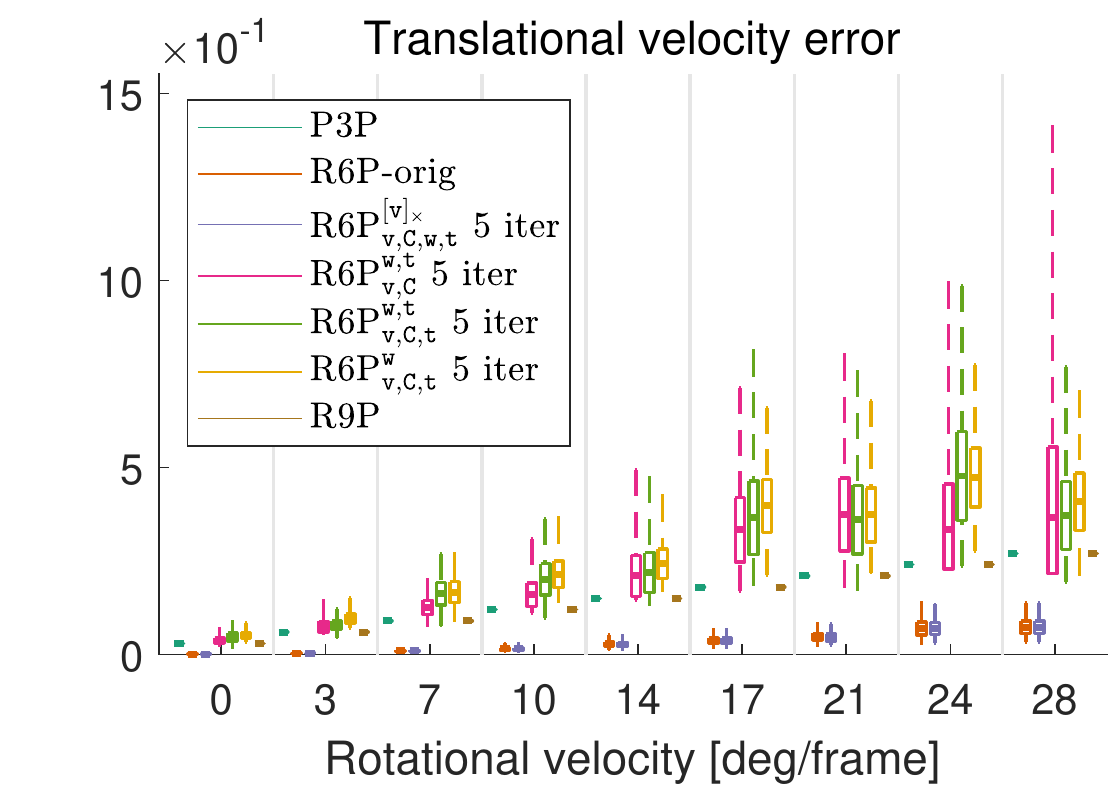}
      \hspace*{0.02\columnwidth}
    \includegraphics[trim={1cm 0 0 0},clip,width=0.45\columnwidth]{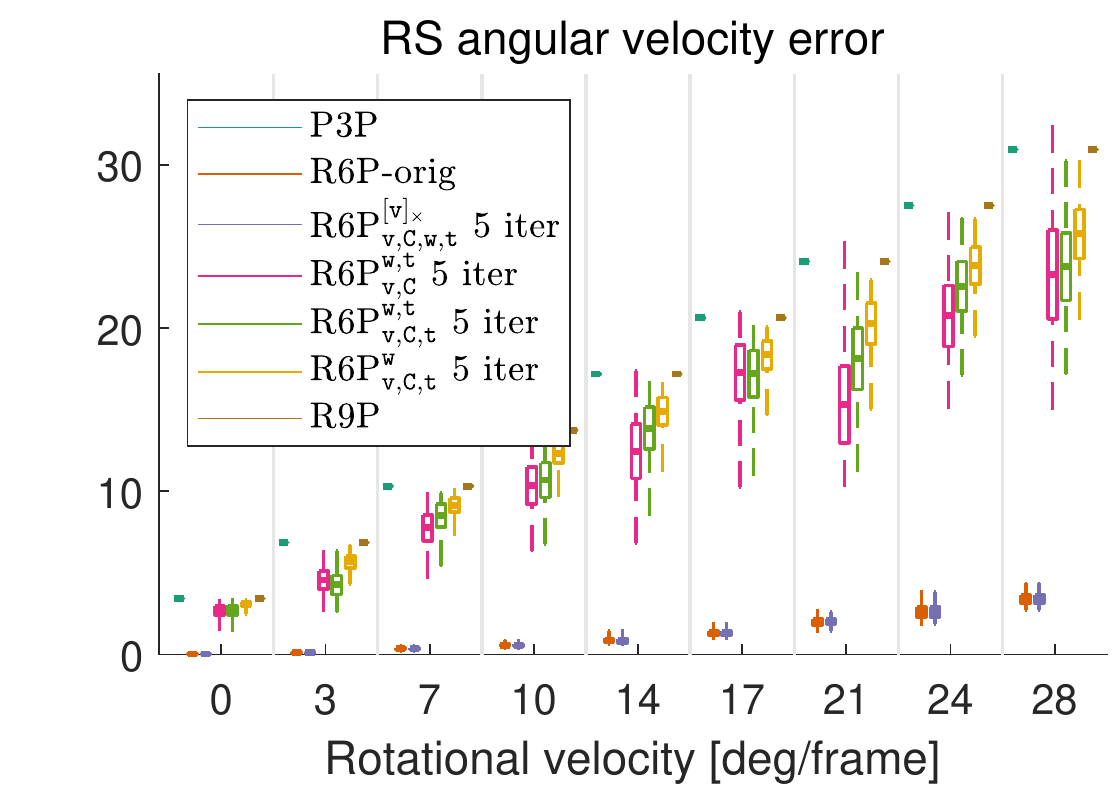}
    \caption{Experiment on synthetic data focusing on the precision of estimated camera poses and velocities. Notice that the performance of \solverall{} is identical to R6P. In terms of camera pose these two solvers are slightly outperformed by R9P. Other linear solvers perform very poorly in all respects.}
    \label{fig:synth_inc_w_t_pose}
\end{figure}

In the first experiment, we gradually increased the camera velocities during capture. The maximum translational velocity was 0.3 per frame and the maximum angular velocity was 30 degrees per frame.
Figure~\ref{fig:synth_inc_w_t_pose} shows the results, from which we can see how the increasing RS deformation affects the estimated camera pose and also estimated camera velocities in those solvers.

\begin{figure}[t]
    \centering
    \includegraphics[width=0.40\columnwidth]{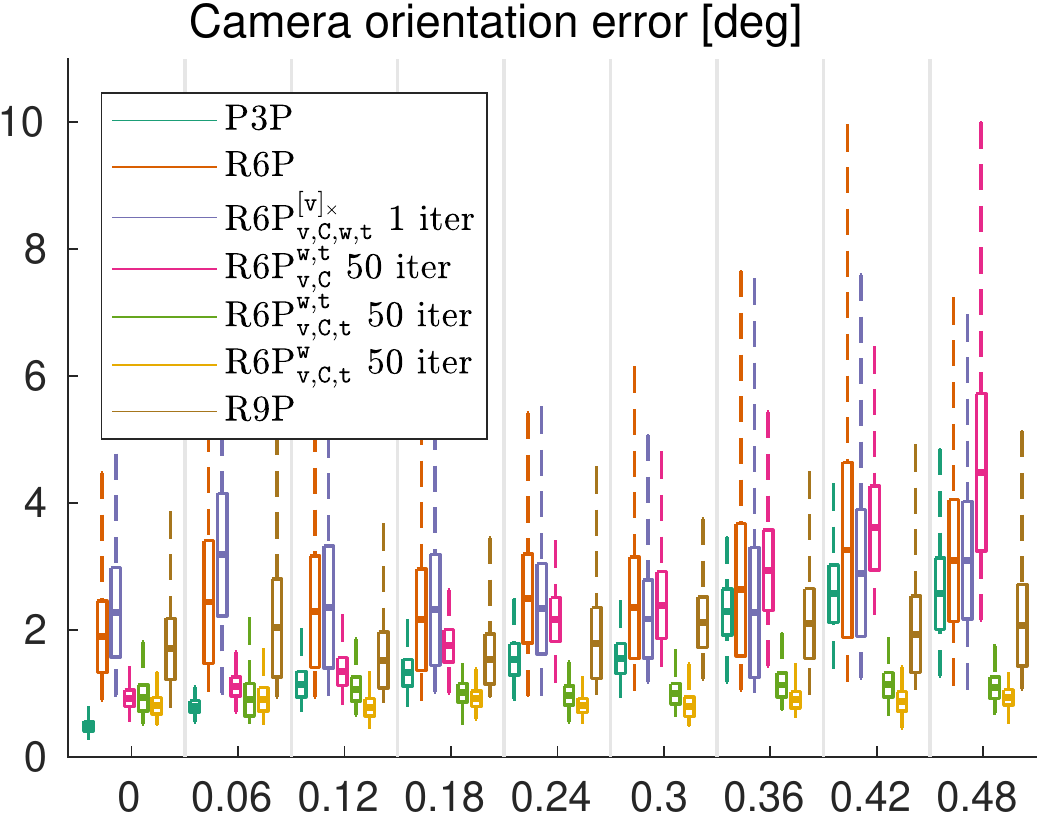}
     \includegraphics[width=0.45\columnwidth]{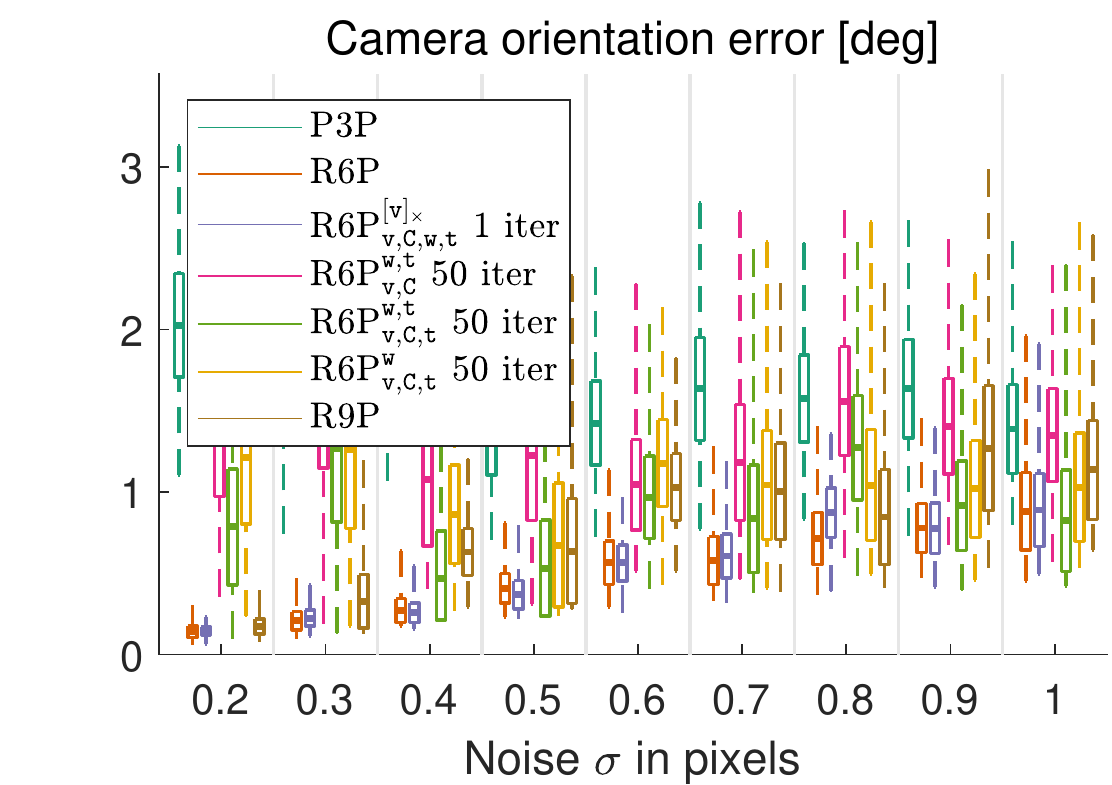}
    \setlength{\belowcaptionskip}{-4pt} 
    \vspace*{-0.5\baselineskip}
    \caption{(Left) Purely translational camera motion, increasing on the x axis. Image noise with $\sigma$ 1pix. Notice that~\solverw{} and~\solverwt{} now outperform all the others. (Right) Performance on general camera motion with increasing image noise.}
    \label{fig:exp_tr_only}
\end{figure}

\begin{figure}[t]
    \centering
    \includegraphics[width=0.4\columnwidth]{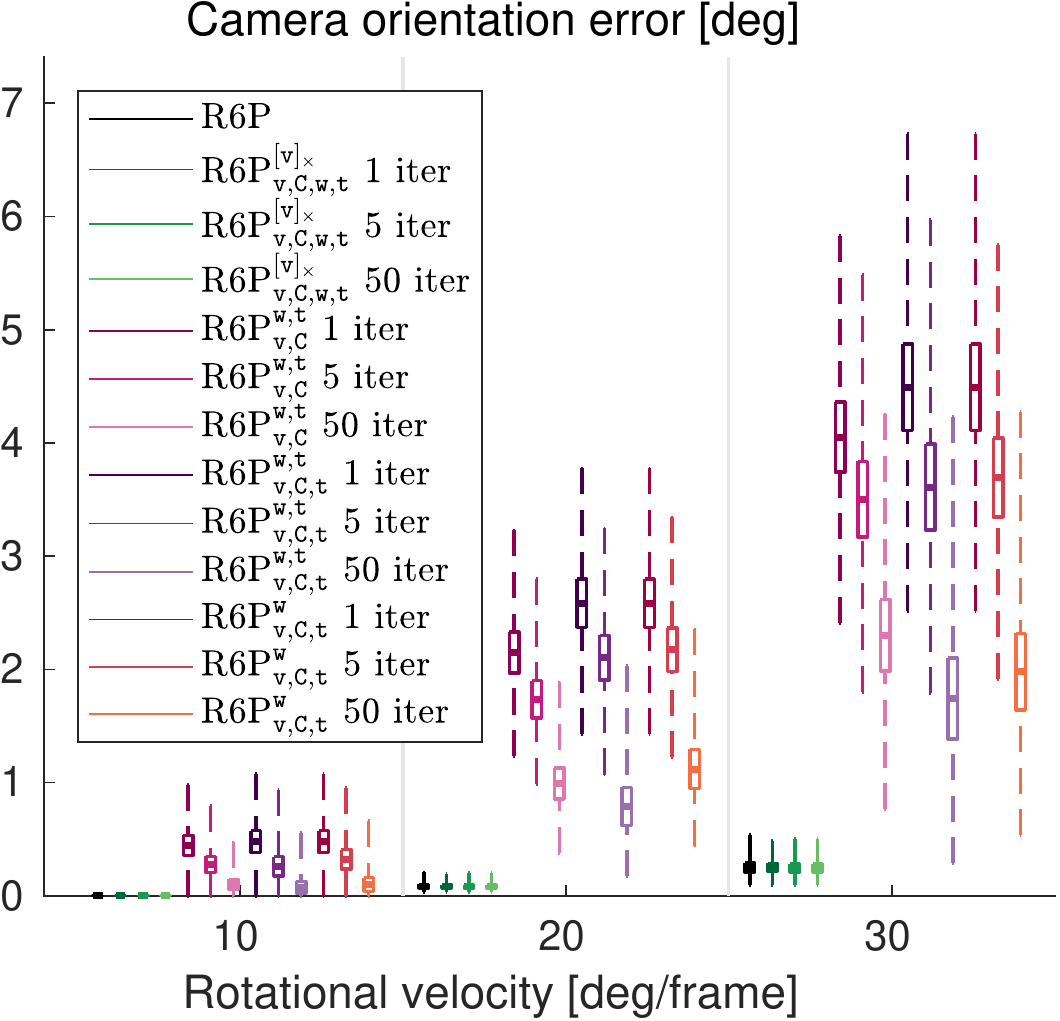}
    \includegraphics[width=0.54\columnwidth]{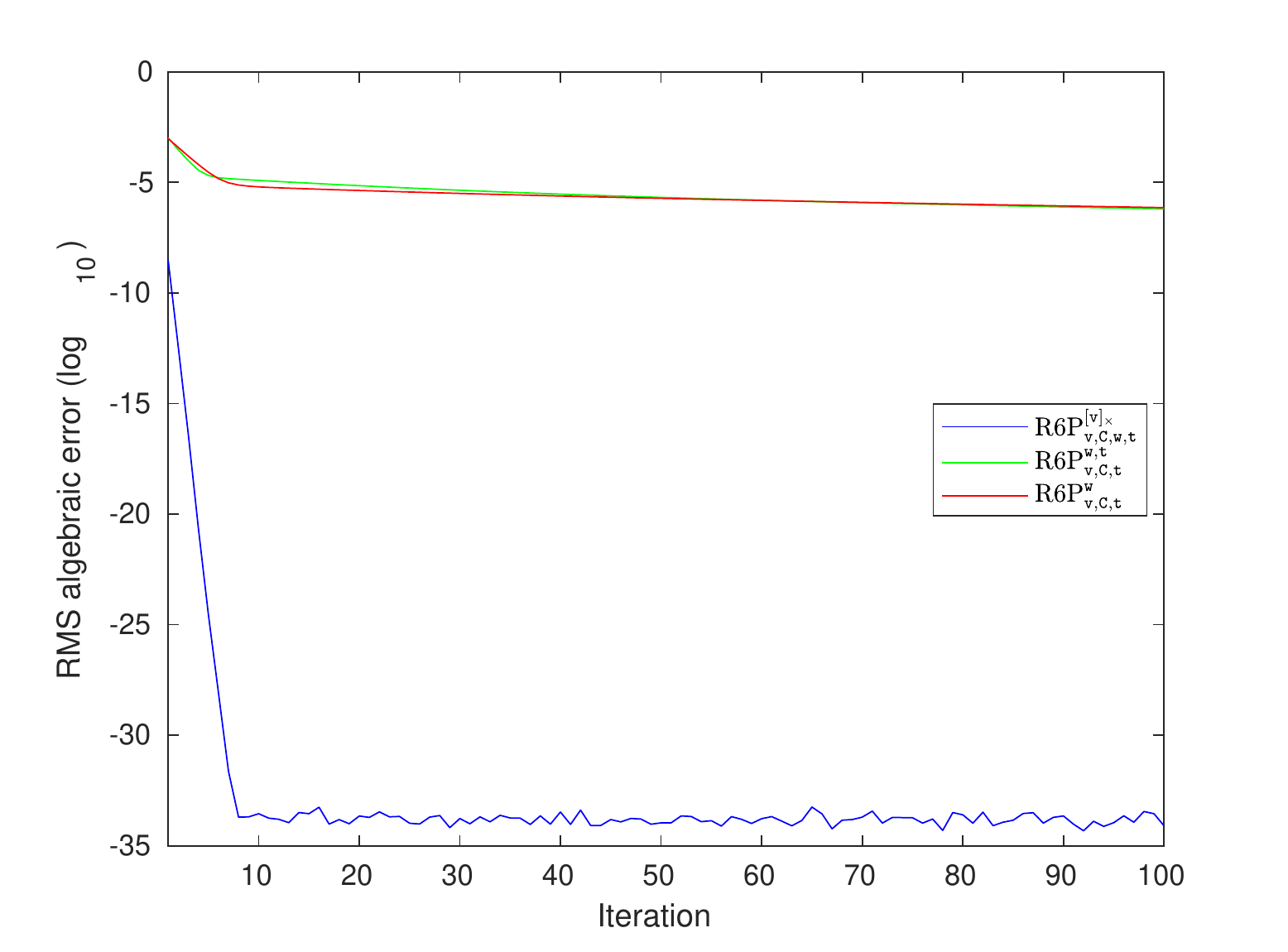}
      \setlength{\belowcaptionskip}{-6pt} 
    \caption{Testing the convergence of the iterative solvers. All iterative solvers have been run with 1, 5 and 50 iterations on data with $\M{R}=\M{I}$ and increasing RS effect (left). Convergence of the algebraic error using the three viable iterative solvers (right).}
    \label{fig:synth_convergence}
\end{figure}

\begin{figure}[t]
    \centering
    \includegraphics[width=0.45\columnwidth]{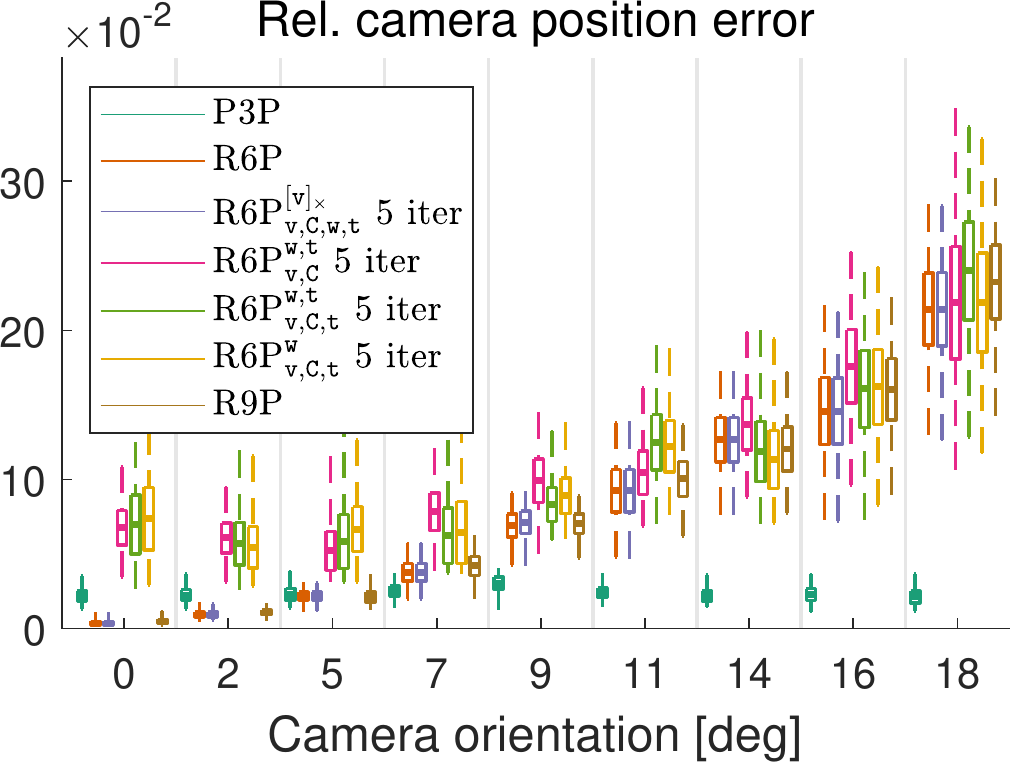}
    \hspace*{0.02\columnwidth}
    \includegraphics[width=0.45\columnwidth]{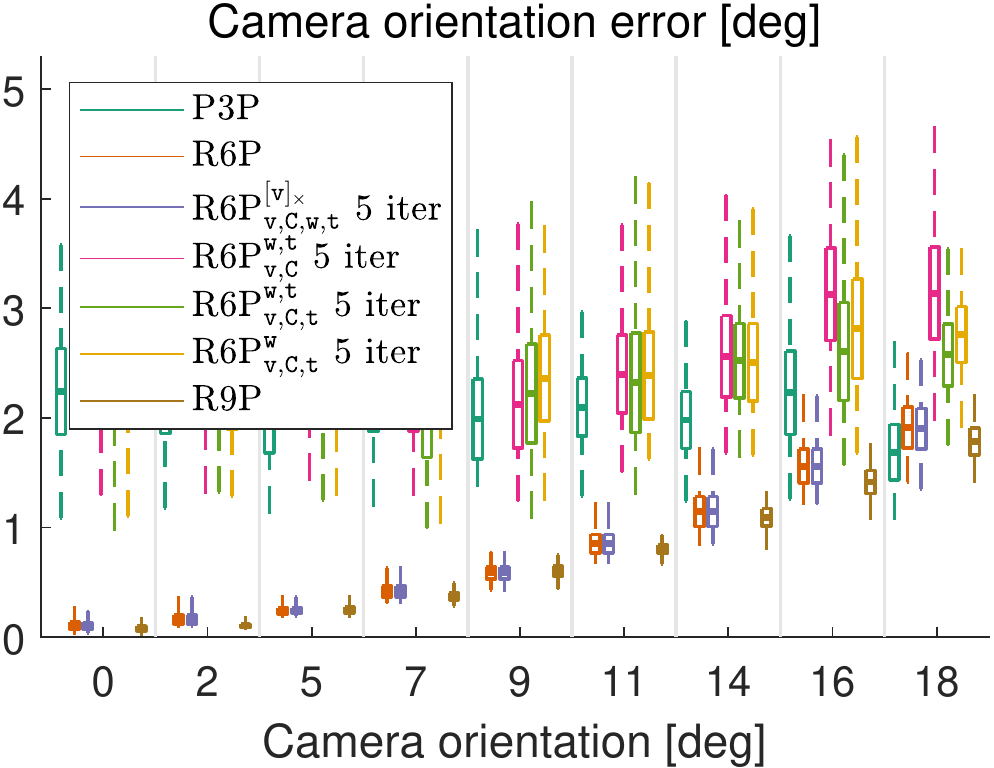} 
    \vspace*{-0.5\baselineskip}
    \caption{Experiment showing the effect of the linearized camera pose which is present in all models. The further the camera orientation is from the linearization point, the worse are the results. \solverall{} matches the results of R6P and so does R9P.}
    \label{fig:synth_inc_R}
\end{figure}

\paragraph{\rm {\bf Rotational and translational motion:}
In agreement with~\cite{Albl-CVPR-2015}, R6P provides much better results than P3P thanks to the RS camera model. The newly proposed solver \solverall{} provides almost identical results to R6P at much lower computation cost (cf. Table~\ref{tab:timings}). The best estimates of the camera pose are provided by R9P at the cost of using more than minimal number of points. The other 6-point iterative solutions are performing really bad, often providing worse results than P3P. 
In the next experiment we tested the sensitivity of the proposed solvers to increasing levels of image noise. Figure~\ref{fig:exp_tr_only} right shows that the new solvers have approximately the same noise sensitivity as R6P~\cite{Albl-CVPR-2015}.
}
\paragraph{\rm {\bf Translational motion only:}
The advantage of solvers \solverwt{} and \solverw{} is when the motion of the camera is purely translational, or close to it, which is a common scenario in, e.g., a moving car or a moving train. In such cases, both original R6P and \solverall{} provide significantly worse estimates of the camera pose. We explain this by the fact that \solverwt{} and \solverw{} are constrained to estimate only camera translation in the initial step, whereas R6P and \solverall{} try to explain the image noise by the camera rotation. See Figure~\ref{fig:exp_tr_only} left. This fact can be used to create a ''joined solver'' that runs both \solverw{} and \solverall{} and gives better performance than R6P~\cite{Albl-CVPR-2015} while still being significantly faster.}
\paragraph{\rm {\bf Convergence:}
For \solverCv{}, \solverw{}, and \solverwt{}, the maximum 5 iterations might not be enough to converge to a good solution, whereas \solverall{} seems to perform at its best. We thus increased the maximum number of iterations. 
Figure~\ref{fig:synth_convergence} (left) shows that the performance of \solverCv{}, \solverw{}, and \solverwt{} is improved by increasing the maximum number of iterations to 50. However, it is still far below the performance of R6P, \solverall{}, and R9P. \solverall{} performs as well as the R6P even with a single iteration, making it two orders of magnitude faster alternative. The algebraic error, evaluated on the equations~(\ref{eq:model_double_lin}), of the three viable solvers converges within 8 stpdf on average, see Figure~\ref{fig:synth_convergence} (right).
}
\paragraph{\rm {\bf The effect of linearized camera rotation model:}
Since all the proposed solvers have a linearized form of the camera orientation, in the same way as R6P~\cite{Albl-CVPR-2015}, we tested how being further from the linearization point affects the performance (Fig.~\ref{fig:synth_inc_R}). The camera orientation was set to be at a certain angle from $\M{R}=\M{I}$.  The camera velocities were set to 0.15 per frame for the translation and 15 degrees per frame for the rotation. In~\cite{Albl-CVPR-2015} the authors show that R6P outperforms P3P in terms of camera center estimation up to 6 degrees away from the initial $\M{R}$ estimate and up to 15 degrees away from $\M{R}$ for the camera orientation estimate. Our results in Figure~\ref{fig:synth_inc_R} show similar behavior and identical results of R6P and \solverall{}. R9P performs comparable to both, even slightly outperforming them in terms of camera orientation estimation. 
}

\begin{figure}[tb]
    \centering
    \includegraphics[width=0.45\columnwidth]{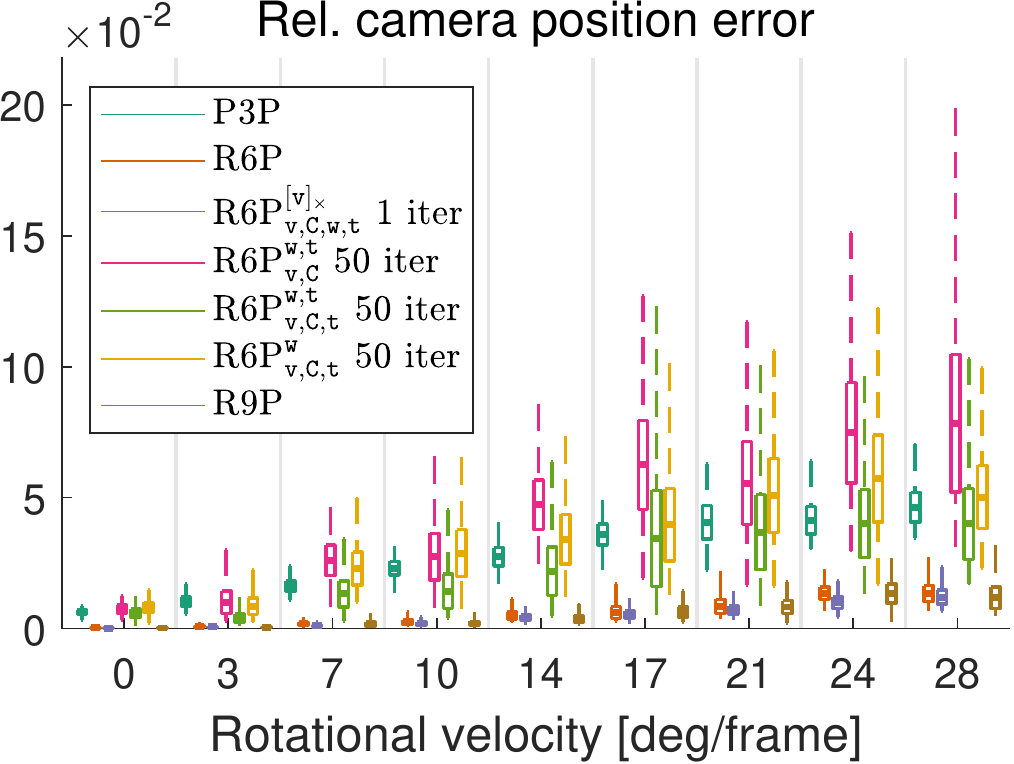}
    \hspace*{0.02\columnwidth}
    \includegraphics[trim={1cm 0 0 0},clip,width=0.45\columnwidth]{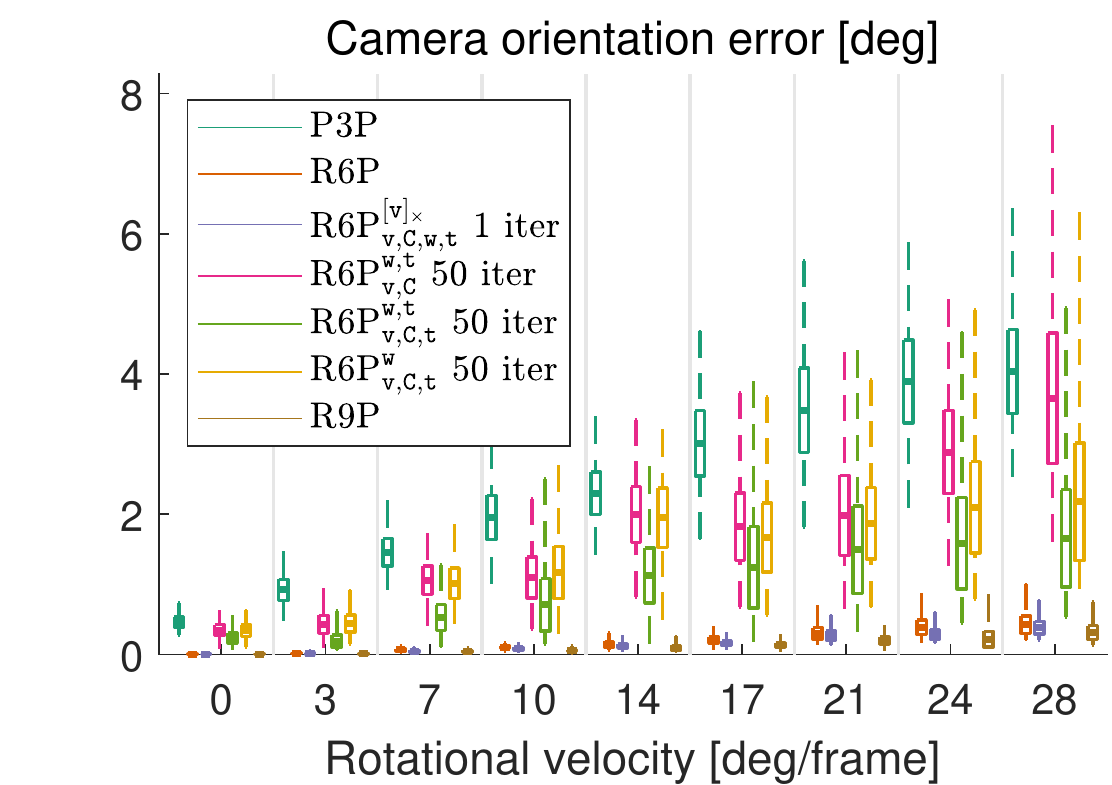} 
    \setlength{\belowcaptionskip}{-4pt} 
    \caption{Increasing the camera motion and estimating camera pose with all solvers being initialized with P3P. \solverall{} and R9P now provide consistently excellent results, comparable or outperforming those of R6P at a fraction of the computation cost. 
    \solverCv{}, \solverw{} and \solverwt{} with 50 iterations now perform better than P3P, but still not as good as the other RS solvers.
    }
    \label{fig:synth_inc_R_p3p_init}
\end{figure}

\begin{paragraph}{\bf Using P3P as initial estimate:}
Last synthetic experiment shows the performance of the solvers when using the initial estimate of $\M{R}$ from the result of P3P. The camera orientation was randomly generated and the camera motion was increased as in the first experiment. P3P was computed first and the 3D scene was pre-rotated using $\M{R}$ from P3P. This shows probably the most practical usage among all R6P solvers. To make the figure more informative, we chose the number of iterations for \solverCv{}, \solverw{}, and \solverwt{} to be 50 as the 5 iterations already proved to be insufficient, see Figure~\ref{fig:synth_inc_w_t_pose}. We also set the maximum number of iterations for \solverall{} to 1, to demonstrate the potential of this solver.
\end{paragraph}

As seen in Figure~\ref{fig:synth_inc_R_p3p_init}, \solverall{} provides at least as good, or even better, results than R6P after only a single iteration. This is a significant achievement since the computational cost of \solverall{} is two orders of magnitude less than of R6P. With 50 iterations the other iterative solvers perform better than P3P, but considering the computational cost of 50 iterations, which is even higher than that of a R6P, we cannot recommend using them in such a scenario.


\begin{table}[tb]
    \centering
    \caption{Average timings on 2.5GHz i7 CPU per iteration for all used solvers.} 
      \label{tab:timings}
    \begin{tabular}{|c|c|c|c|c|c|c|c|}
         \hline
         solver & P3P & R6P & \solverall{} & \solverw{} & \solverwt{} & \solverCv & R9P \\ \hline
         time per iteration & $3\mu s$ & $1700 \mu s$ & $10 \mu s$ & $24 \mu s$ & $30 \mu s$ & $27 \mu s$ & $20 \mu s$ \\ \hline
         max $\#$ of solutions & 4 & 20 & 1 & 1 & 1 & 1 & 1 \\ \hline
    \end{tabular}
\end{table}
\paragraph{\rm {\bf Computation time:}
The computation times for all the tested solvers are shown in Table~\ref{tab:timings}. One iteration of \solverall{} is two orders of magnitude faster than R6P. According to the experiments, even one iteration of \solverall{} provides very good results, comparable with R6P and 5 iterations always match the results of R6P or even outperform them at 34$\times$ the speed. Note that R9P can be even faster than \solverall{} because it is non-iterative and runs only once and is therefore as fast as 2 iterations of \solverall{}.  
One iteration of \solverCv{}, \solverw{} and \solverwt{} is around three times slower than \solverall{} but still almost two orders of magnitude faster than R6P. }


\subsection{Real data}
\label{sec:exp_real}
We used the publicly available datasets from~\cite{Hedborg2012} and we show the results of the same frames shown in~\cite{Albl-CVPR-2015} (seq1, seq8, seq20 and seq22) in order to make a relevant comparison. We also added one more real dataset (House), containing high RS effects from a fast moving drone carrying a GoPro camera. The 3D-2D correspondences were obtained in the same way as in~\cite{Albl-CVPR-2015} by reconstructing the scene using global shutter images and then matching the 2D features from the RS images to the reconstructed 3D points.

We performed RANSAC with 1000 iterations for each solver to estimate the camera pose and calculated the number of inliers. The inlier threshold was set to 2 pixels in the case of the data from~\cite{Hedborg2012} which was captured by handheld iPhone at 720p and to 8 pixels for the GoPro footage which was recorder in 1080p. The higher threshold in the second case allowed to capture a reasonable number of inliers even for such fast camera motions.  
\begin{figure}
    \centering
    \includegraphics[width = 0.48\columnwidth]{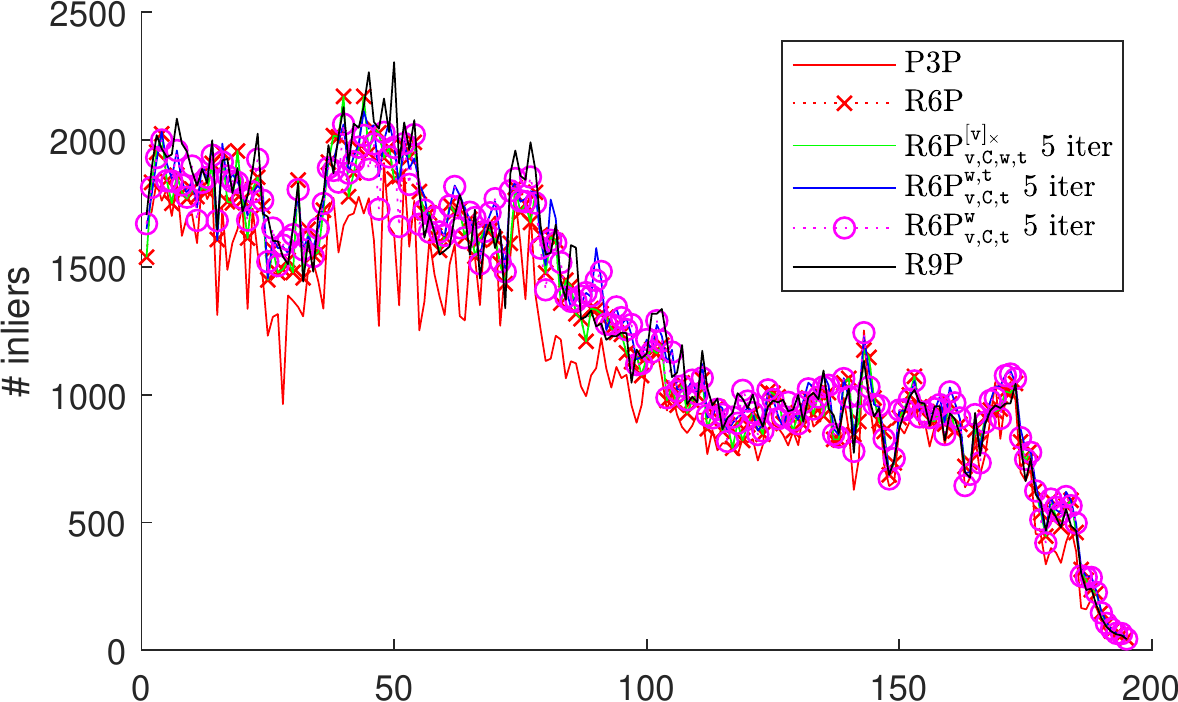}
    \includegraphics[width = 0.48\columnwidth]{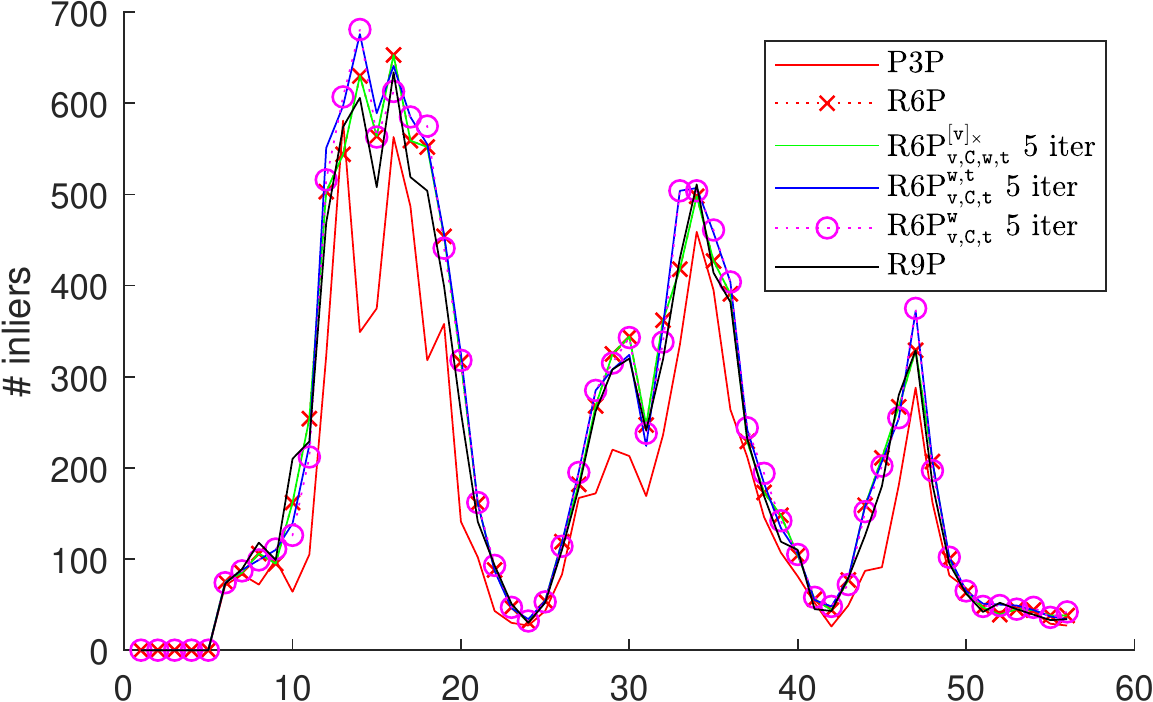}\\
    \includegraphics[width = 0.48\columnwidth]{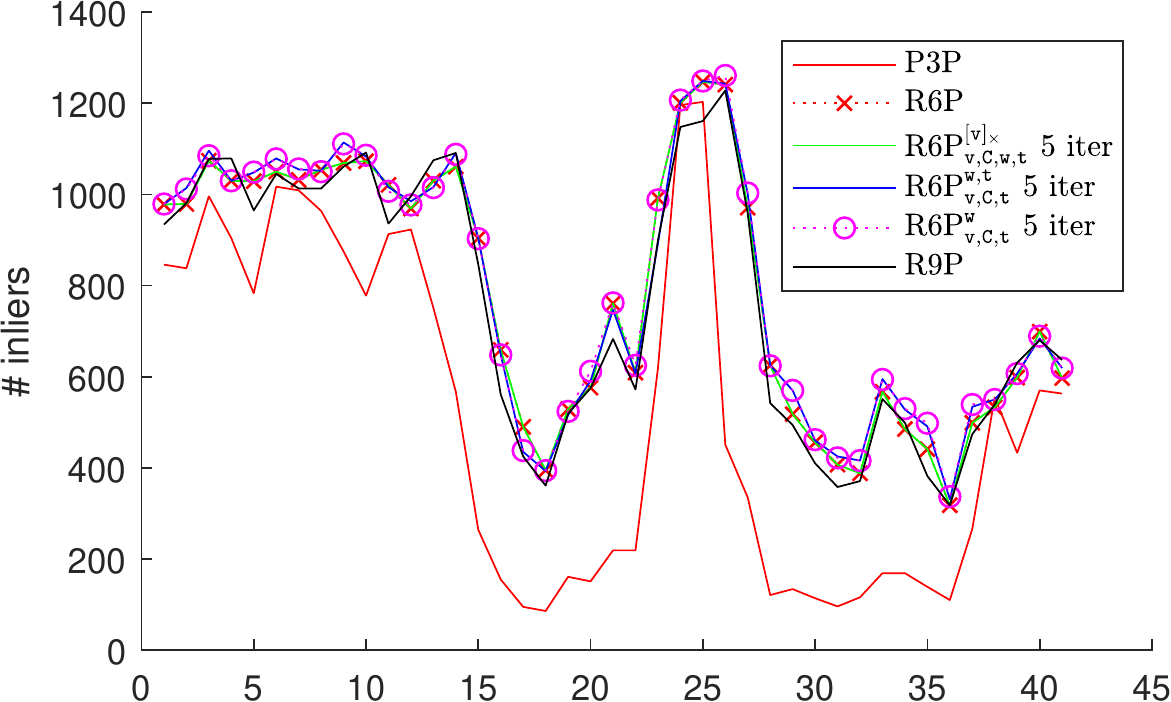}
    \includegraphics[width = 0.48\columnwidth]{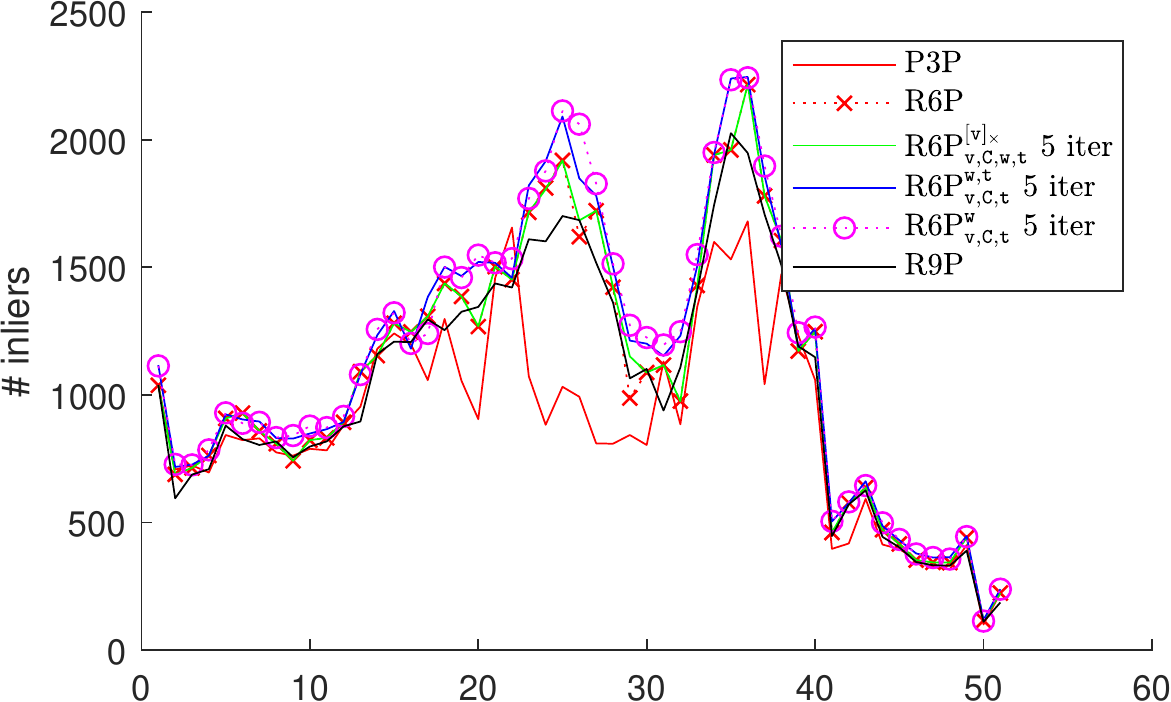}\\
    \includegraphics[width = 0.48\columnwidth]{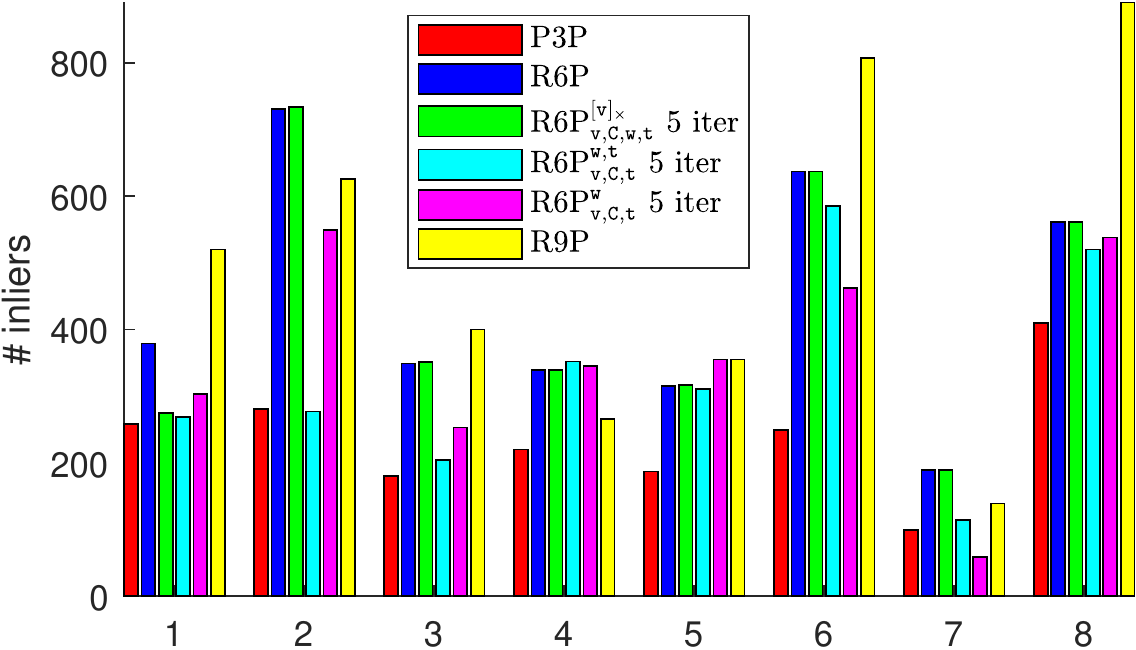}
    \includegraphics[width = 0.48\columnwidth]{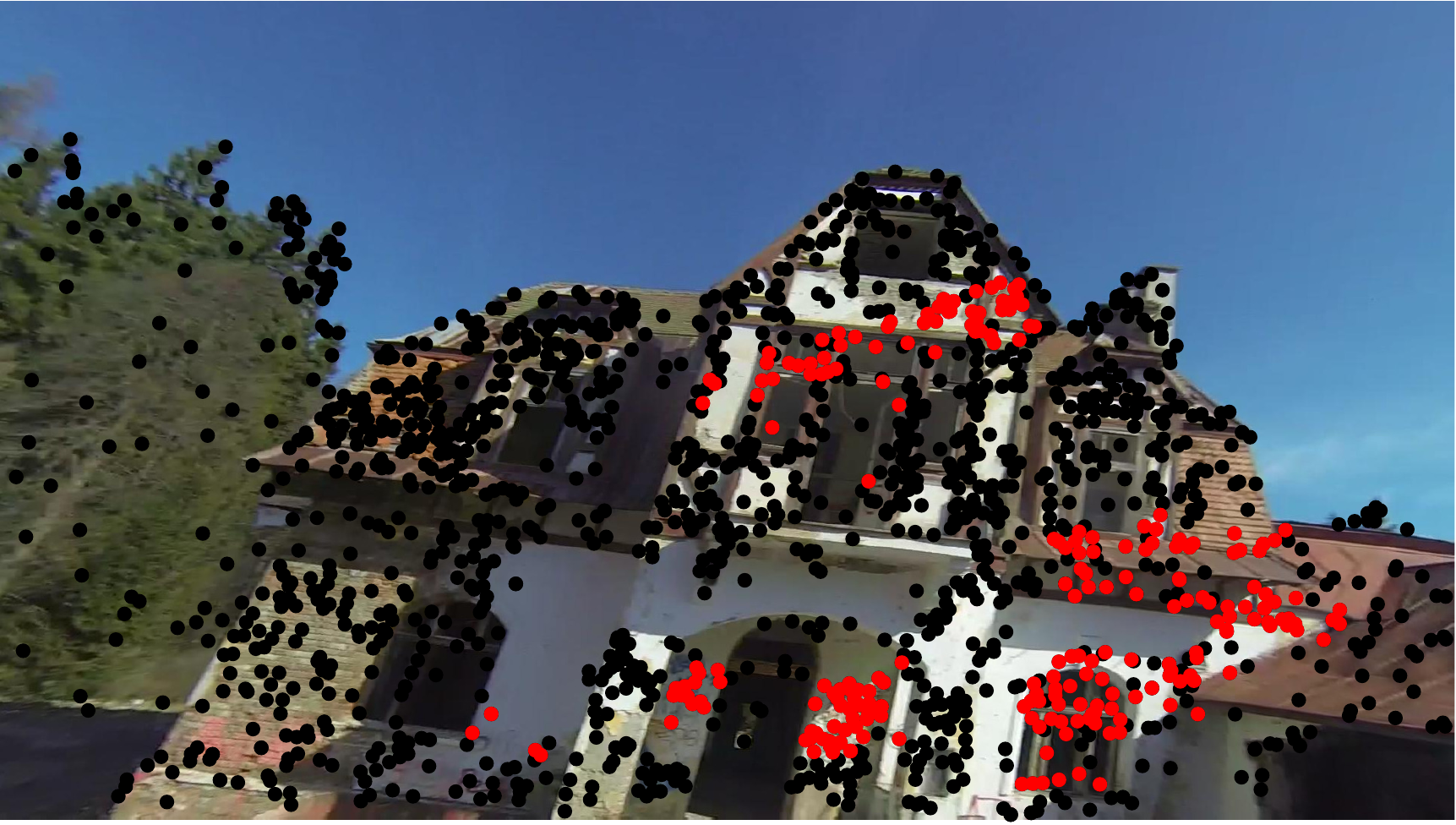}\\
    \includegraphics[width = 0.48\columnwidth]{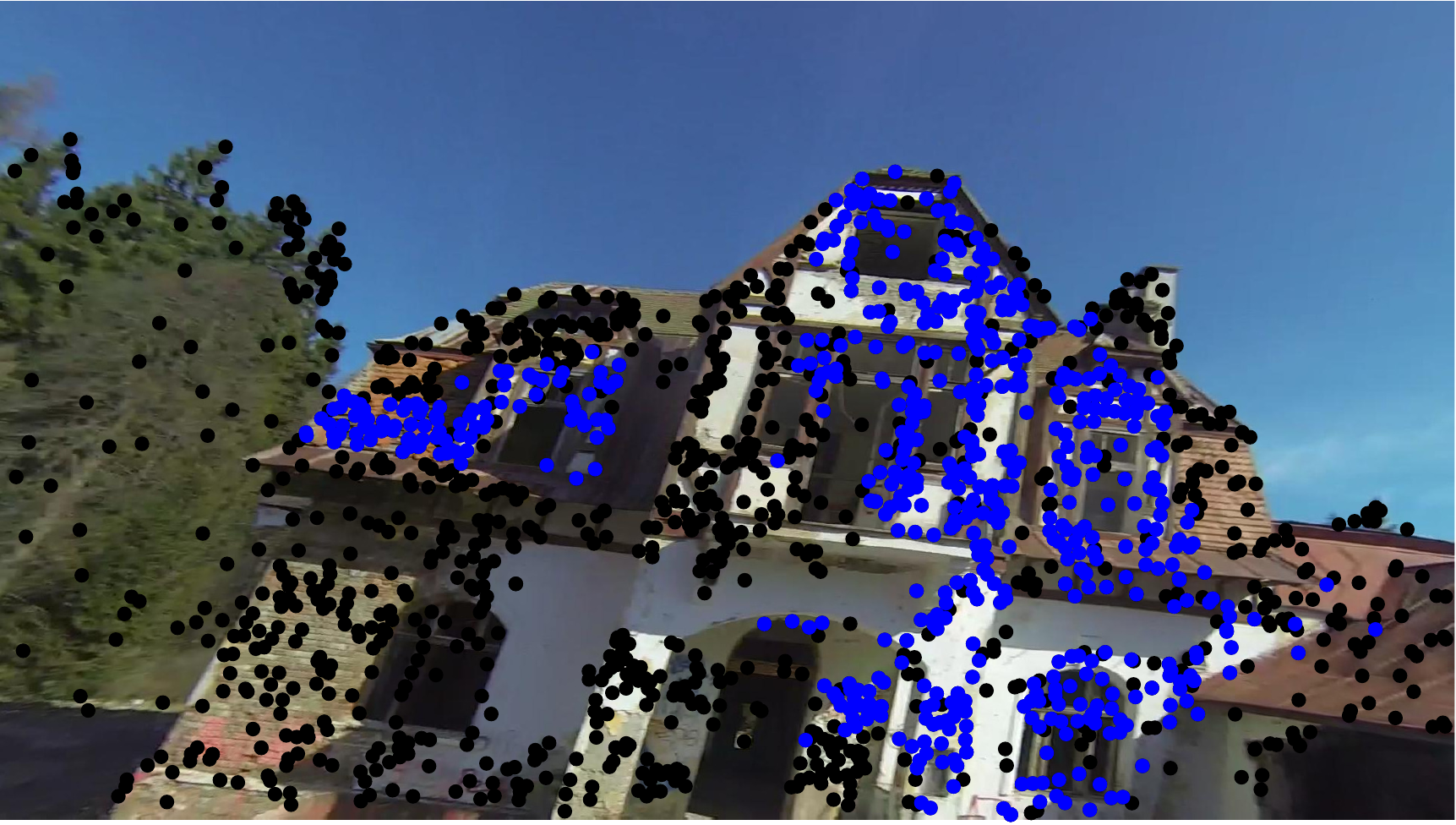}
    \includegraphics[width = 0.48\columnwidth]{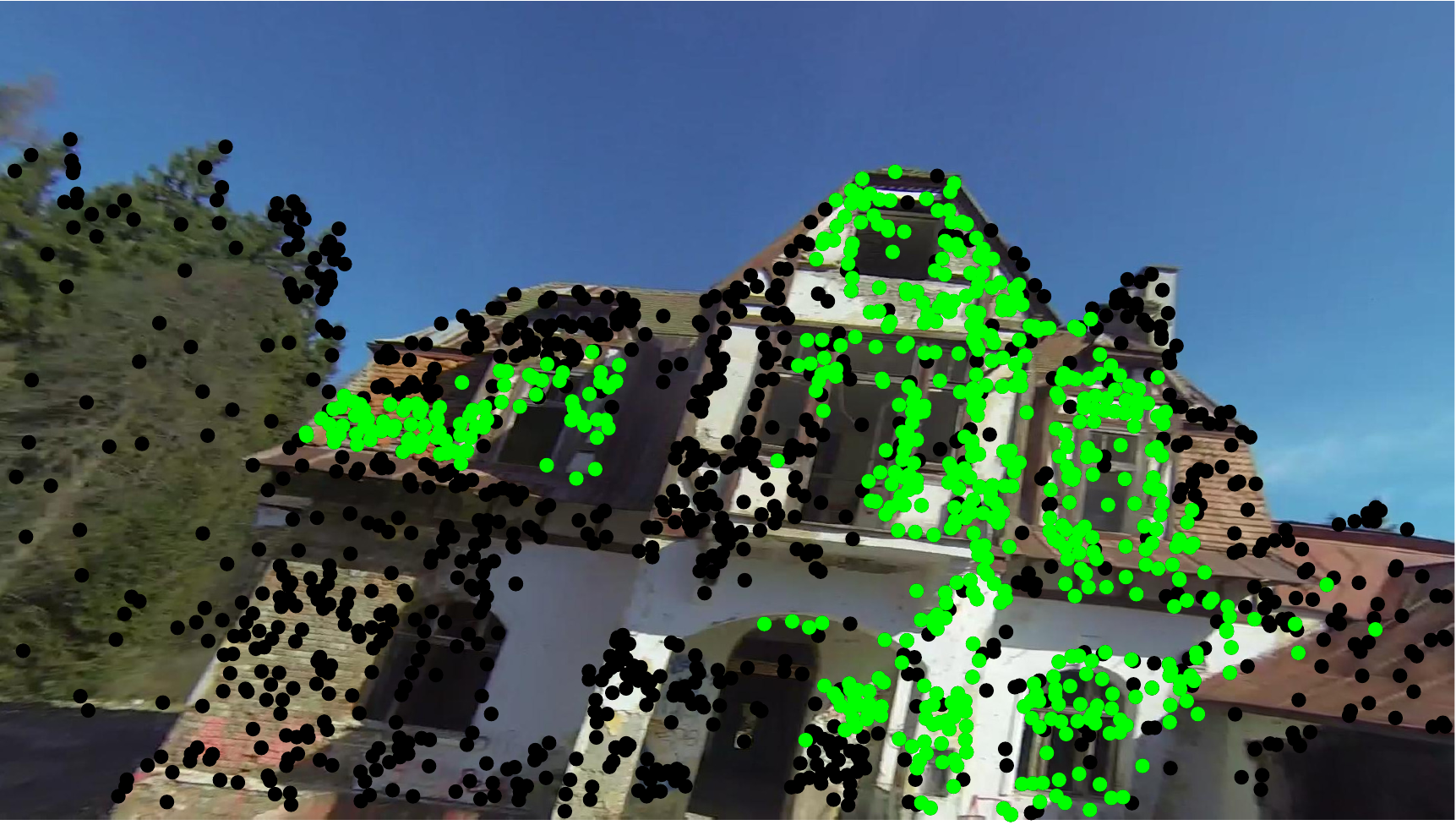}\\
    \includegraphics[width = 0.48\columnwidth]{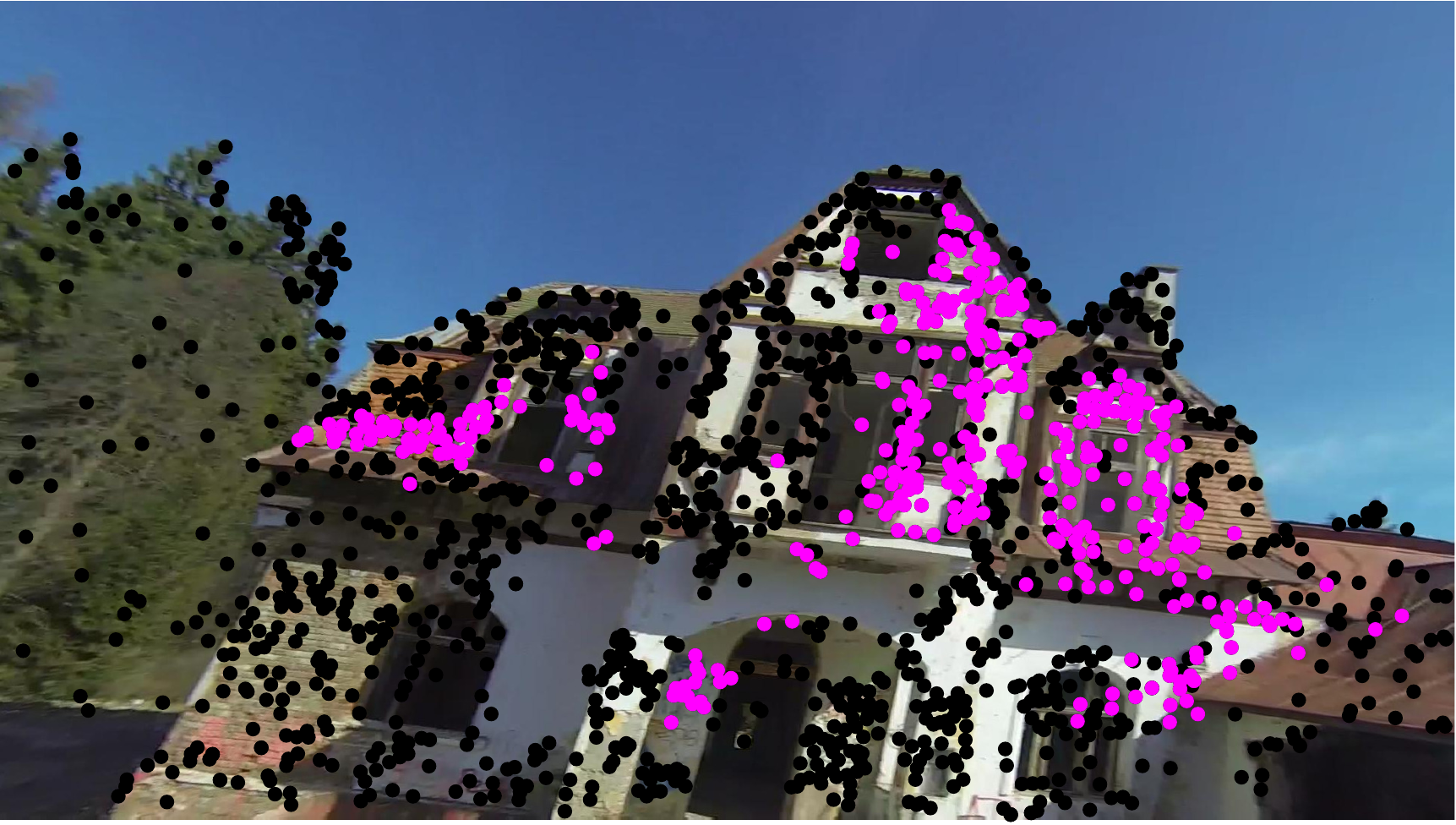}
    \includegraphics[width = 0.48\columnwidth]{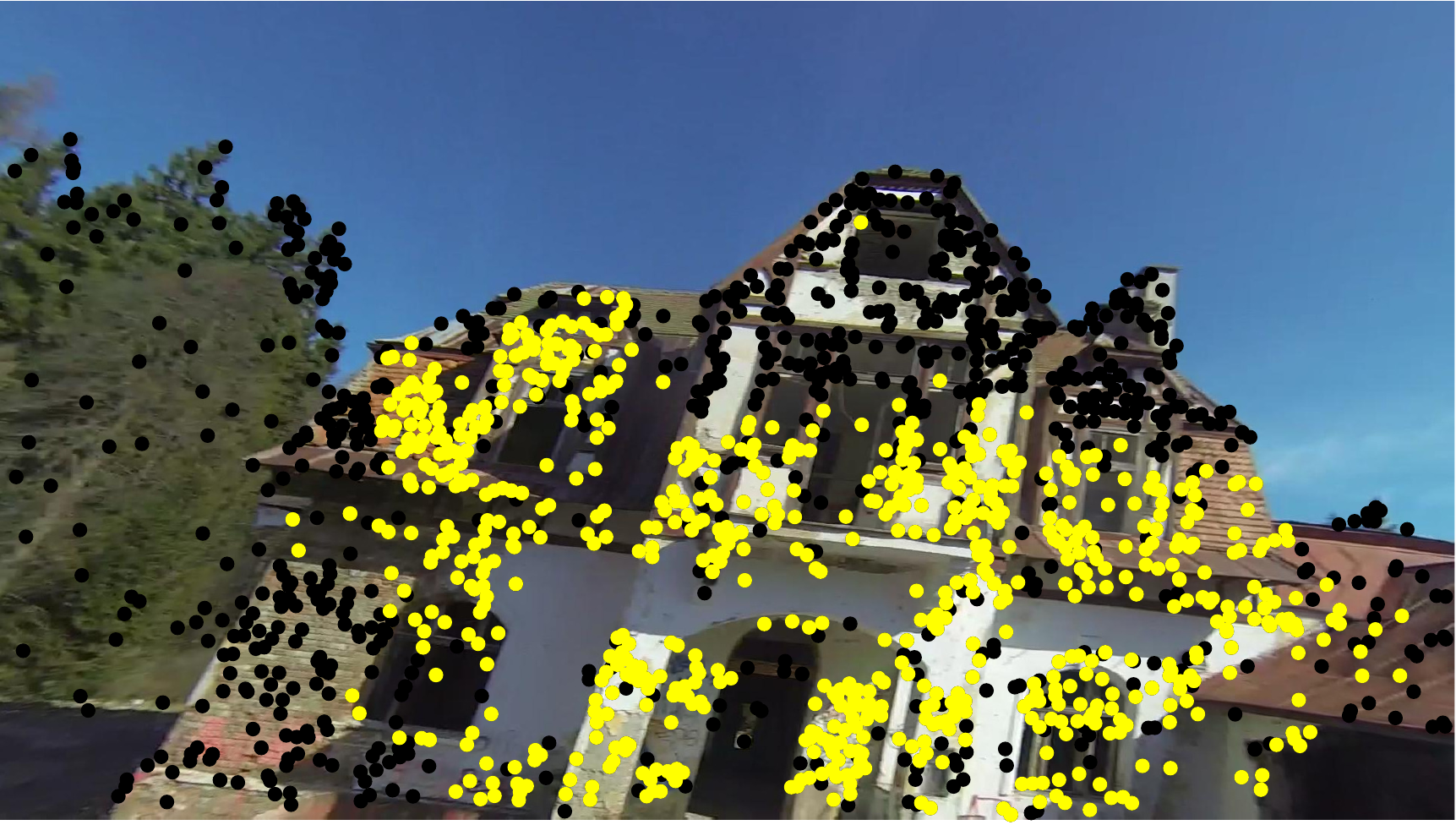}\\
    \caption{Number of inliers on real data sequences. From top to bottom, left to right: seq01, seq08, seq20, seq22 and House. The x axis contains frame numbers. The bar graph for the House figure is used because there is no temporal relationship between adjacent frames so a line graph does not make sense. Following are sample images from the House dataset frame 6, containing a high amount of RS distortion. In this frame, R9P provided significantly more inliers than other methods. The results of R6P and \solverall{} are again identical, with the small exception of the first frame. The colored inliers in the sample images follow the same colors of algorithms as in the bar graph for House sequence.}
    \label{fig:real_inliers}
\end{figure}
The results in Figure~\ref{fig:real_inliers} show the number of inliers captures over the sequences of images. We see that the performance of \solverall{} with 5 iterations is virtually identical to R6P. The results of \solverwt{} and \solverw{} are also very similar and often outperform R6P and \solverall{}, except for the most challenging images in the House dataset. 

The performance of \solverCv{} is unstable, sometimes performing comparable to or below P3P. In seq20 in particular, there is almost exclusively a fast translational camera motion. The drop in performance can therefore be explained by \solverCv{} being the only solver that does not estimate the translational velocity $\V{t}$ in the first step. R9P performs solidly across all the experiments and on the most challenging House dataset it even provides significantly better results.

\begin{figure}
    \centering
    \includegraphics[width = 0.35\columnwidth]{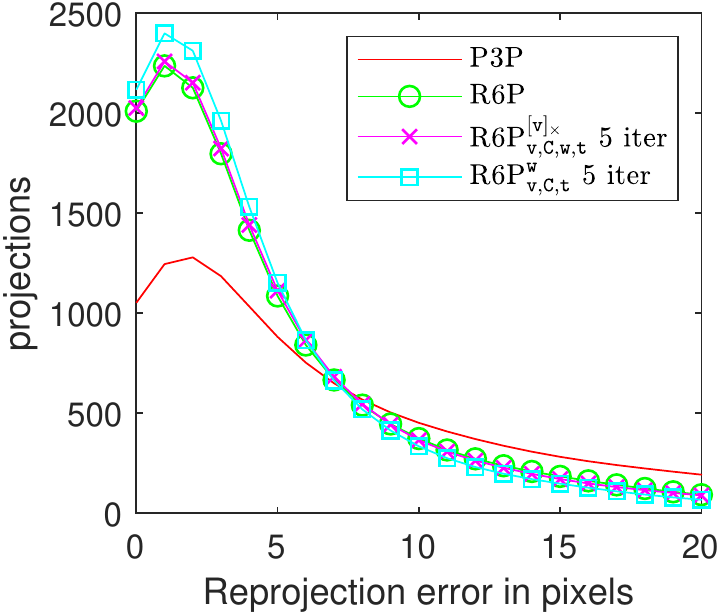}
    \includegraphics[height=0.295\columnwidth]{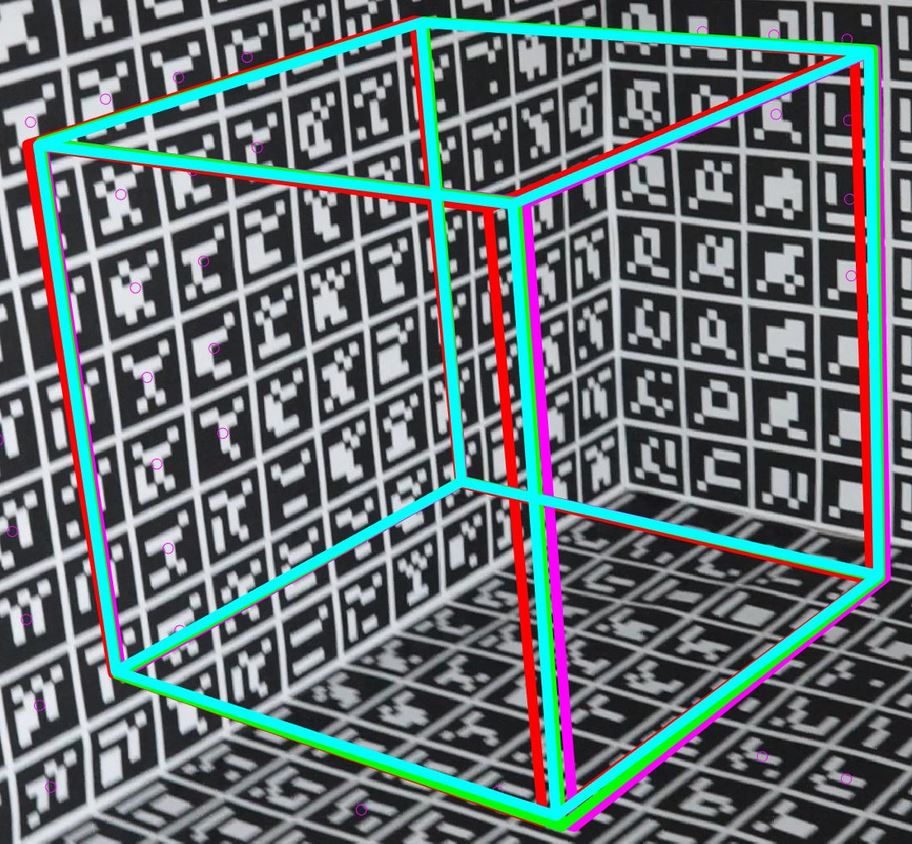} 
    \includegraphics[height=0.295\columnwidth]{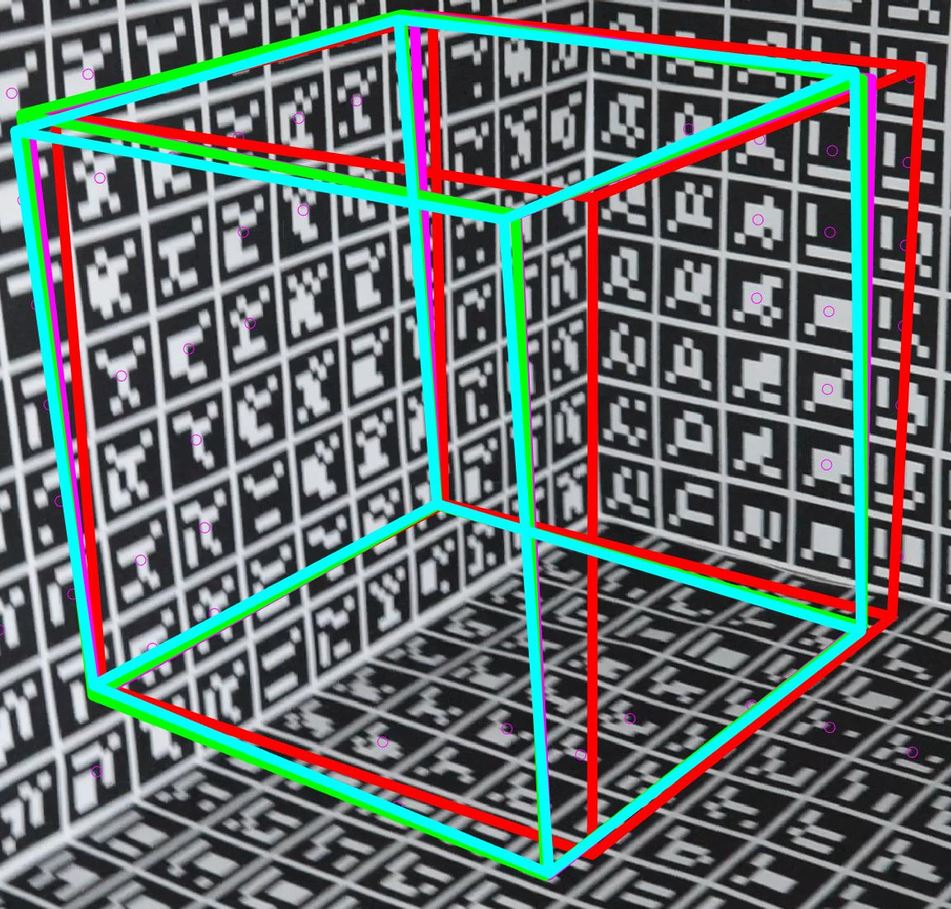} 
    \caption{Histogram of reprojection errors on the Aruco markers in the augmented reality experiment.  The rolling shutter absolute pose solvers (R6P in magenta,\solverall{} in green, \solverw{} in cyan) keep the cube in place during camera motion whereas P3P (red) reprojects the cube all over the place.}
    \label{fig:real_aruco_reproj}
\end{figure}

To test another useful case of camera absolute pose, which is augmented reality, we created an environment filled with Aruco~\cite{aruco} markers in known positions. We set up the markers in such a way that they covered three perpendicular walls. The scene was recorded with a camera performing translational and rotational motion, similar to what a human does when looking around or shaking the head.


All solvers were used in RANSAC with 100 iterations to allow some robustness to outliers and noise. Note that 100 iterations of RANSAC would take at least 200ms for R6P excluding the inlier verification. That makes R6P not valuable for real time purposes (in practice only less than 10 iterations of R6P would give realtime performance). On the other hand, 100 runs of \solverall{} with 5 iterations take around 5ms (200fps) and \solverw{} takes around 12.5ms (80fps). 
We did not test solvers \solverCv{}, \solverwt{} and R9P in this experiment.
This is because the performance of \solverCv{} is unstable, the performance of \solverwt{} is almost identical to \solverw{} and 
with R9P we do not have a way to extract the camera motion parameters and the reprojection without these parameters does not provide fair comparison.  

We evaluated the reprojection error in each frame on all the detected markers. The results are shown in Figure~\ref{fig:real_aruco_reproj}. All the rolling shutter solvers outperform P3P in terms of precision of the reprojections. \solverall{} again provides identical performance to R6P. \solverw{} has a slight edge over the others, which is interesting, considering its poor performance on the synthetic data.

Figure~\ref{fig:real_aruco_reproj} gives a visualization of the estimated camera pose by reprojecting a cube in front of the camera. There is a significant misalignment between the cube and the scene during camera motion when using P3P pose estimate. In comparison, all the rolling shutter solvers keep the cube much more consistent with respect to the scene. 
\section{Conclusions}
We revisited the problem of rolling shutter camera absolute pose and proposed several new practical solutions. The solutions are based on iterative linear solvers that improve the current state-of-the-art methods in terms of speed while providing the same precision or better. The practical benefit of our solvers is also the fact that they provide only a single solution, compared to up to 20 solutions of R6P~\cite{Albl-CVPR-2015}.

The overall best performing \solverall{} solver needs only a single iteration to provide similar performance to R6P while being approximately 170x faster. At 5 iterations the performance of R6P is matched while the new \solverall{} solver is still approximately 34x faster than R6P. This allows for much broader applicability, especially in the area of augmented reality, visual SLAM and other real-time applications. 

We also proposed 3 other iterative linear solvers (\solverCv{}, \solverwt{}, \solverw{}) that alternate between estimating different camera pose and velocity parameters. These three solvers are slower than \solverall{} but still almost two orders of magnitude faster than R6P. While not as precise as R6P or \solverall{} in the synthetic experiments, they proved usefulness on the real data, providing more inliers and better reprojections than P3P and even R6P. We presented these three solvers mainly because they follow the concept of making the rolling shutter absolute pose equations linear by alternatively fixing some variables and then others. Although \solverwt{} and \solverw{} do not offer the fastest and most precise results, they performed best in some of the experiments, especially for purely translational motion, and we think they are worth mentioning.

Last but not least we presented a non-iterative linear solver that uses 9 correspondences. This solver is as fast as 2 iterations of \solverall{} and proved to be the most precise in terms of estimated camera pose in the synthetic experiments and provided solid performance on the real data.

Altogether, this paper presents a big step forward in practical computation of rolling shutter camera absolute pose, making it more available in real world applications.

\bibliographystyle{splncs}

\end{document}